\documentclass[pdflatex,sn-mathphys-num]{sn-jnl}

\usepackage{graphicx}%
\usepackage{multirow}%
\usepackage{amsmath,amssymb,amsfonts}%
\usepackage{amsthm}%
\usepackage{mathrsfs}%
\usepackage[title]{appendix}%
\usepackage{xcolor}%
\usepackage{textcomp}%
\usepackage{manyfoot}%
\usepackage{booktabs}%
\usepackage{algorithm}%
\usepackage{algorithmicx}%
\usepackage{algpseudocode}%
\usepackage{listings}%
\usepackage{caption}

\usepackage{enumerate}

\usepackage{hyperref}
\usepackage{titlesec}
\usepackage{appendix}
\titleformat{\paragraph}[block]{\normalfont\normalsize\bfseries}{}{0em}{}
\titlespacing*{\paragraph}{0pt}{1ex}{0pt} 
\hypersetup{hypertex=true,
	colorlinks=true,
	linkcolor=blue,
	anchorcolor=blue,
	citecolor=blue}

\usepackage{array}%
\usepackage{diagbox}%
\usepackage{epstopdf}%

\usepackage{pdflscape}

\usepackage{makecell}%
\usepackage{longtable}%
\usepackage{threeparttable}%

\usepackage{float}%
\usepackage{bm}
\usepackage[section]{placeins}

\theoremstyle{thmstyleone}%
\theoremstyle{thmstyletwo}%

\theoremstyle{thmstylethree}%

\begin{document}

\title[Article Title]{Impact of large language models on peer review opinions from a fine-grained perspective: Evidence from top conference proceedings in AI}


\author[1]{\fnm{Wenqing} \sur{Wu}}\email{winchywwq@njust.edu.cn}

\author*[1]{\fnm{Chengzhi} \sur{Zhang}}\email{zhangcz@njust.edu.cn}

\author[2]{\fnm{Yi} \sur{Zhao}}\email{yizhao93@ahu.edu.cn}

\author[1]{\fnm{Tong} \sur{Bao}}\email{tbao@njust.edu.cn}

\affil[1]{\orgdiv{Department of Information Management}, \orgname{Nanjing University of Science and Technology}, \orgaddress{\city{Nanjing}, \postcode{210094}, \state{Jiangsu}, \country{China}}}
\affil[2]{\orgdiv{School of Management}, \orgname{Anhui University}, \orgaddress{\city{Hefei}, \postcode{230601}, \state{Anhui}, \country{China}}}


\abstract{With the rapid advancement of Large Language Models (LLMs), the academic community has faced unprecedented disruptions, particularly in the realm of academic communication. The primary function of peer review is improving the quality of academic manuscripts, such as clarity, originality and other evaluation aspects. Although prior studies suggest that LLMs are beginning to influence peer review, it remains unclear whether they are altering its core evaluative functions. Moreover, the extent to which LLMs affect the linguistic form, evaluative focus, and recommendation-related signals of peer-review reports has yet to be systematically examined. In this study, we examine the changes in peer review reports for academic articles following the emergence of LLMs, emphasizing variations at fine-grained level. Specifically, we investigate linguistic features such as the length and complexity of words and sentences in review comments, while also automatically annotating the evaluation aspects of individual review sentences. We also use a maximum likelihood estimation method, previously established, to identify review reports that potentially have modified or generated by LLMs. Finally, we assess the impact of evaluation aspects mentioned in LLM-assisted review reports on the informativeness of recommendation for paper decision-making. The results indicate that following the emergence of LLMs, peer review texts have become longer and more fluent, with increased emphasis on summaries and surface-level clarity, as well as more standardized linguistic patterns, particularly reviewers with lower confidence score. At the same time, attention to deeper evaluative dimensions, such as originality, replicability, and nuanced critical reasoning, has declined. These phenomena are more obvious when comparing LLM-assisted and non-LLM-assisted reviews, and the aspects mentioned in LLM-assisted reports have a modest positive influence on informativeness of the recommendations.}

\keywords{Peer review, Large language model, Academic communication, LLM-assisted text detection, Fine-grained analysis}



\maketitle

\section{Introduction} \label{sec1}
\noindent Peer review is a critical quality control mechanism in the academic research and publication process \cite{r1}. Its primary purpose is to ensure the rigor and credibility of academic research, assist authors in improving their work, and identify potential errors and shortcomings \cite{r2,r3}. However, in recent years, the peer review mechanism has faced widespread criticism due to the surge in paper submissions and the shortage of domain experts qualified to serve as reviewers \cite{r4,r5}, particularly at top artificial intelligence (AI) conferences \cite{r6}. Current peer review processes face several challenges, including bias \cite{r49}, variability in review quality \cite{r49}, unclear reviewer motivations \cite{r50}, and imperfect review mechanism \cite{r51}. As submission volumes continue to rise, these issues are becoming increasingly pronounced.  Some researchers have sought to mitigate these problems by enhancing fairness \cite{r50}, reducing biases among novice reviewers \cite{r49}, calibrating noisy peer review ratings \cite{r52}, and improving mechanisms for matching papers with reviewers' expertise \cite{r53,r54}. Other studies \cite{r7,r8,r9,r10,r11,r12,r31} have explored the use of natural language processing techniques to support or refine the peer review process. These studies introduce the possibility of leveraging artificial intelligence to assist overburdened scientists in the peer review process \cite{r13,r14}. While these technologies may aid reviewers to some extent, their impact on the peer review process still requires further study. \\
\indent In recent years, the impressive capabilities demonstrated by large language models (LLMs) \cite{r15} have sparked extensive research and discussion within the academic community. At the same time, concerns \cite{r16,r17} have emerged within the academic community about the potential erosion of peer review by LLMs. Researchers have also begun to study and analyze the application and impact of LLMs in the peer review process. For example, Liang et al. \cite{r18} not only evaluated the effectiveness of GPT-4 in generating scientific feedback but also proposed \cite{r19} a method to estimate the extent of LLMs usage in peer review texts. They found that some reviews from recent AI conferences may have been modified by LLMs. Latona et al. \cite{r20} investigated the prevalence and impact of LLM-assisted peer reviews at the ICLR 2024, finding that LLM-assisted reviews significantly influence review scores and submission acceptance rates. While these preliminary studies indicate that LLMs have begun to affect peer review, whether they are altering the core functions of peer review remains underexplored. As LLMs become increasingly integrated into scholarly workflows, it is therefore important to analyse their impacts on peer review from multiple perspectives, including linguistic patterns, evaluation aspects, and the influence of reviewer recommendations. Understanding these effects can help refine peer review practices, ensure fairness and transparency, and provide actionable insights for adapting to the evolving academic landscape. Notably, major conferences such as NeurIPS\footnote{\href{https://neurips.cc/Conferences/2024/ReviewerGuidelines}{https://neurips.cc/Conferences/2024/ReviewerGuidelines}. Since 2018, NIPS has been renamed to NeurIPS. Therefore, this paper uniformly uses NeurIPS to refer to the conference.} have not yet established explicit policies on whether reviewers may use LLMs to assist in writing their reports. This policy ambiguity underscores the importance of examining how LLM assistance may already be shaping the linguistic and evaluative characteristics of peer review texts.\\
\indent This study is guided by the following three research questions:\\
\indent \textbf{RQ1:} How has the emergence of LLMs affected the linguistic complexity and aspect-level content expression of peer review texts?\\
\indent \textbf{RQ2:} In comparison to non-LLM-assisted reviews, which evaluation aspects are more prominently emphasized in LLM-assisted reviews?\\
\indent \textbf{RQ3:} How do the evaluation aspects emphasized in LLM-assisted reviews relate to reviewers' scoring and confidence levels?\\
\indent In summary, we conduct a fine-grained analysis of peer review reports before and after the emergence of LLMs, aiming to address the three research questions outlined above.
To answer Research Question 1 (RQ1), we examine the linguistic characteristics of review texts, such as average word and sentence length, lexical sophistication, syntactic complexity, and the distribution of evaluation aspects, to determine whether and how linguistic expression has evolved with the introduction of LLM assistance.
To address Research Question 2 (RQ2), we automatically annotate review sentences with predefined evaluation aspects (e.g., clarity, summary, and soundness) and compare their distributions between LLM-assisted and non LLM-assisted reviews, in order to identify which aspects are more prominently emphasized by LLM-assisted reviewers.
For Research Question 3 (RQ3), we analyse the relationship between reviewers? recommendations and the identified evaluation aspects, assessing how LLM assistance influences the connection between reviewers' focal aspects and their recommendations.
Finally, to complement these analyses, we apply a maximum likelihood estimation model trained on both expert-authored and AI-generated reference texts to estimate review texts that may have been substantially modified or generated by LLMs.\\
\indent The main contributions of this paper are as follows:\\
\indent First, this paper conducts a fine-grained statistical analysis of the review report texts from two artificial intelligence conferences. We find that the proliferation of LLMs has had the most pronounced impact on ICLR review reports, as evidenced by an observable increase in word and sentence length, along with a significant rise in the use of nominal subjects. Additionally, the average length of reviews from reviewers with a confidence score of 1 has shown an upward trend in recent years.\\
\indent Second, we analyzed the proportion of evaluation aspects in the review reports of LLM-assisted and non-LLM-assisted reviewers. We found that compared to non-LLM-assisted reviewers, LLM-assisted reviewers significantly increased the proportion of summaries, while the proportion of evaluations focusing on originality decreased.\\
\indent Finally, we explored the relationship between the evaluation aspects mentioned in the review reports of LLM-assisted reviewers and the scores they assigned. The results indicate that the impact of various evaluation aspects on scoring is relatively weak, and the correlation with reviewers' confidence scores is generally low.\\
\indent The code and dataset for this paper can be accessed at \href{https://github.com/njust-winchy/LLM_impact.}{https://github.com\\/njust-winchy/LLM\_impact}    
\section{Related work}
\subsection{Pre-LLM Approaches to Automated Peer Review}
\label{subsec2.1}
Automated Scholarly Paper Review (ASPR) \cite{r21} refers to the process in which computers or intelligent machines independently evaluate the content of a scholarly paper and generate a review report automatically. Currently, research on ASPR is still in its early stages, with machines only serving as an aid to human reviewers. For ASPR, the availability of large datasets is crucial. As pioneers in this field, Kang et al. \cite{r22} introduced PeerRead, a publicly available dataset for research purposes, consisting of 14.7K paper drafts and corresponding peer review reports from top venues, including ACL, NeurIPS, and ICLR. On this basis, Wang et al. \cite{r23} presents ReviewRobot, a novel tool designed to assist human reviewers by automatically assigning review scores and generating constructive comments across multiple categories, demonstrating high accuracy and effectiveness in enhancing the peer review process. Li et al. \cite{r24} introduced a multi-task shared structure encoding approach for predicting peer-review aspect scores of academic papers, demonstrating improved performance over single-task and naive multi-task methods by effectively leveraging auxiliary task information. Furthermore, Yuan et al. \cite{r8} proposed ASAP-Review, a dataset annotated with evaluation aspects of review content, and trained a targeted summarization model for generating peer reviews. Later, Yuan and Liu \cite{r9} proposed an end-to-end knowledge-guided review generation framework for scientific papers, introducing an oracle pre-training strategy to enhance the model's understanding and coverage of review aspects. These studies represent early-stage auxiliary methods for automated academic peer review that laid the groundwork for further advances prior to the emergence of LLMs.\\


\subsection{LLMs for Scientific Peer Review: Applications and Impacts}
\label{subsec2.2}
The emergence of LLMs has sparked a global wave of research, including investigations into their application in peer review and their impact on the peer review process \cite{r55,r56,r57}. Liang et al. \cite{r18} first evaluated GPT-4's utility in generating scientific feedback, revealing that GPT-4's feedback overlaps significantly with human peer reviews and is perceived as helpful by researchers, suggesting its potential to enhance the peer-review process despite noted limitations. Robertson \cite{r26} conducted a pilot study on the use of GPT-4 in the peer review process, demonstrating that LLM-generated reviews can be as useful as human reviews and highlighting the potential applications for addressing resource constraints in peer review. Liu and Shah \cite{r27} demonstrated promising results in using LLMs to identify errors and verify checklist tasks in peer reviews, although limitations were noted in abstract quality comparisons. Furthermore, Thelwall \cite{r28} evaluated GPT-4's capabilities in journal article evaluations and found its accuracy insufficient for full automation, recommending editorial oversight. Zhou et al. \cite{r29} assessed GPT-3.5 and GPT-4 via score prediction and review generation, highlighting valuable insights but also significant limitations. Du et al. \cite{r30}  analyzed LLM-generated reviews compared to human-written reviews, noting the LLMs' potential to identify deficiencies.\\
\indent Amidst the flourishing development of LLMs, research has begun focusing on optimization strategies for LLM-generated peer reviews. Jin et al. \cite{r31} introduce an LLM-based peer review simulation framework AGENTREVIEW that addresses the complexities and privacy concerns of traditional peer review analysis, revealing significant biases in paper decisions. Gao et al. \cite{r32} presented REVIEWER2, a two-stage review generation framework that explicitly models review aspect distributions, supported by a dataset of 27,000 papers and 99,000 annotated reviews. Tan et al. \cite{r33} reformulated peer review as a multi-turn dialogue among authors, reviewers, and decision-makers, with novel evaluation metrics and a large-scale dataset. D'Arcy et al. \cite{r34}  proposed MARG, a feedback generation framework using multiple LLM instances for internal discussion to enhance feedback specificity. Yu et al. \cite{r35} introduced SEA, an automated reviewing framework aimed at improving the quality and consistency of LLM-generated reviews. Wu et al. \cite{r58} explored the potential of LLMs to enhance post-publication peer review, demonstrating that fine-tuned models can effectively identify high-quality articles but still face challenges in providing consistent and context-sensitive evaluations.\\
\indent Beyond generation and optimization, recent studies have also investigated the real-world impact of LLM-assisted reviewing. Recently, Liang et al. \cite{r19} proposed a method to estimate the proportion of LLM-generated or modified texts in peer review corpora, applying it to AI conference reviews post-ChatGPT release. Similarly, Latona et al. \cite{r20} found that at least 15.8\% of ICLR 2024 reviews were LLM-assisted and that these reviews tended to assign higher scores, with such papers more likely to be accepted. Inspired by these findings, we propose three research questions to conduct a fine-grained comparative analysis of peer-review reports before and after LLMs entered the public sphere (i.e., prior to and following the release of ChatGPT), with a particular focus on changes in linguistic patterns, evaluative aspects, and recommendation-related signals.

\subsection{Detection of Large Language Model-Generated Content}
\label{subsec2.3}
LLMs possess enough linguistic capabilities for academic writing \cite{r36} and other scholarly tasks \cite{r37,r38}, which has inevitably raised concerns within the academic community \cite{r17}. Consequently, efforts to detect the use of LLMs have also begun. Mitchell et al. \cite{r39} proposed DetectGPT, which detects text generated by LLMs by analyzing the negative curvature regions of the model's log probability function. Yang et al. \cite{r40} proposed a novel training-free detection method, Divergent N-Gram Analysis (DNA-GPT), which identifies machine-generated text by analyzing the differences between original and regenerated text segments. Furthermore, Li et al. \cite{r41} developed a comprehensive testbed by collecting texts from diverse human writings and LLM-generated deepfake texts from various LLMs. Liang et al. \cite{r19} proposed a simple and effective method for estimating the proportion of text within a large corpus that has been significantly altered or generated by LLM. \\
\section{Methodology}
\begin{figure*}[h]%
	\centering
	\includegraphics[width=1\textwidth]{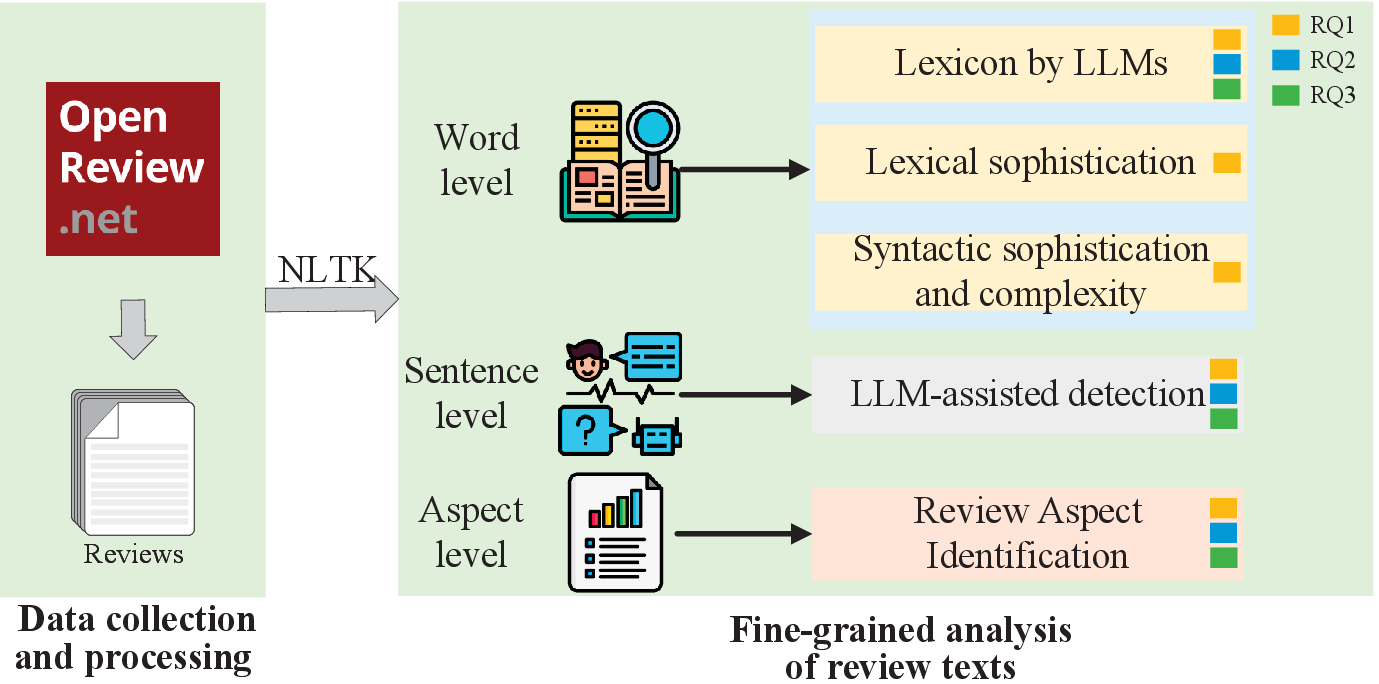}
	
	\caption{\centering{Framework of this study. Colored blocks on the right denote the corresponding research questions.}}
	\label{fig:1}
\end{figure*}

In contrast to the previous studies, we adopted a more refined analytical approach, focusing on the characteristics at the word, sentence, and aspect levels, while Liang et al. \cite{r19} and Latona et al. \cite{r20} primarily conducted analysis at the overall text level. This methodological difference enables our research to provide a deeper understanding and more detailed analysis of the text. It is important to note that this study detects review comments that may have been assisted by LLMs, rather than those entirely generated by LLMs.
Figure \ref{fig:1} presents the research framework for the fine-grained analysis of peer review texts, which primarily includes peer review data collection and processing, and fine-grained analysis of review texts. 
\subsection{Data Collection and Processing}
Considering the availability of large-scale data, this study selects two top-tier conferences in the field of computer science, ICLR and NeurIPS, using OpenReview\footnote{\href{https://openreview.net/}{https://openreview.net/}} as the data sources. It is important to note, however, that while this study uses these two conferences as examples, the proposed methodology is generalizable and can be applied and analyzed in other domains, if data in a similar format is available. The rationale for selecting this field and these two conferences is that they provide years of continuous and publicly available peer review data, which serves as a valuable reference. The detailed data statistics are shown in Table \ref{tab3}. It is important to note that since we rely on publicly available data, and neither OpenReview nor the released datasets include NeurIPS 2020, the data for that year are therefore missing.\\
\begin{table}[h]
	\vspace{-0.8cm}
	\caption{\centering{Number of ICLR and NeurIPS reviews and papers per year.}}\label{tab3}%
	\centering
	\begin{tabular}{@{}m{15mm}<{\centering} m{15mm}<{\centering}m{20mm}<{\centering}m{25mm}<{\centering}m{20mm}<{\centering}m{20mm}<{\centering} @{}}
		\toprule
		\textbf{Venue} & \textbf{Year} &  \textbf{\# Papers} & \textbf{\# Reviews} & \textbf{\# Accept} &\textbf{\# Reject}\\
		\midrule
		\multirow{8}{*}{ICLR}    & 2017 & 489 & 1,498 & 245 & 244\\
		& 2018 &911&	2,748&	425&	486\\
		& 2019 & 1,579&	4,764&	502&	1,077\\
		& 2020 & 2,213&	6,721&	687&	1,526 \\
		& 2021  & 2,972&	11,058&	860&	2,112\\
		& 2022  & 3,325&	12,777&	733&	2,592\\
		& 2023  & 4,915&	18,560&	1,575&	3,340\\
		& 2024  & 5,749&	22,245&	2,261&	3,488\\
		& 2025  &10,497& 42,571& 3,703& 6,794\\
		\hline
		\multirow{7}{*}{NeurIPS}& 2016& 569& 3,183& 569& 0\\
		&2017&	679&	1,941&	679&	0\\
		&2018&	1,009&	3,014&	1,009&	0\\
		&2019&	1,428&	4,253&	1,428&	0\\
		&2021&	2,766&	10,729&	2,630&	136\\
		&2022&	2,824&	10,330&	2,671&	153\\
		&2023&	3,394&	15,171&	3,218&	176\\
		&2024&	4,234&	16,635&	4,033&	201\\
		\hline
		Total&& 49,613&	188,198&	27,228&	22,325\\
		\bottomrule 
	\end{tabular}
\end{table}
\indent These reviews include textual evaluations, confidence scores, and overall score ratings. The confidence score is measured on a consistent scale from 1 to 5, reflecting the reviewers' confidence in the comments they have provided. The overall score is based on a scale ranging from 1 to 10. The descriptive statistics for the two types of scores are provided in Appendix \ref{app2}, Tables \ref{tab:taba2} and \ref{tab:taba3}. Each score is accompanied by a detailed definition to clarify its meaning and provide guidance, the detail can be found in the Reviewer Guidelines\footnote{\href{https://iclr.cc/Conferences/2024/ReviewerGuide}{https://iclr.cc/Conferences/2024/ReviewerGuide},\\ \href{https://neurips.cc/Conferences/2024/ReviewerGuidelines}{https://neurips.cc/Conferences/2024/ReviewerGuidelines}}, which outline the expectations and criteria for reviewing papers. These guidelines include instructions for evaluating various aspects of the submitted content, such as clarity, originality, soundness, and significance, and require scoring for certain aspects. Then, we use the Natural Language Toolkit (NLTK\footnote{\href{https://www.nltk.org}{https://www.nltk.org}}) to segment the text from peer review reports into words and sentences, facilitating subsequent aspect identification and LLM-assisted detection. 

\subsection{Review Sentence Aspect Identification}
\label{sec3.2}
To conduct a fine-grained analysis of peer review texts, it is necessary to identify evaluation aspects within the segmented peer review sentences. Based on the review guidelines and the study by Yuan et al. \cite{r8}, we can define eight review aspects: summary, motivation, originality, soundness, substance, replicability, meaningful comparison, and clarity. Descriptions of these evaluation aspects can be found at \href{https://github.com/neulab/ReviewAdvisor/blob/main/materials/AnnotationGuideline.pdf}{https://github.com/neulab/ReviewAdvisor/blob/main/materials/AnnotationGuideli\\ne.pdf}. \\
\indent For sentence-level evaluation aspect identification, we adopted the pre-trained aspect identification model from the study by Yuan et al. \cite{r8} and further optimized it. Their model was trained on manually annotated data with the addition of heuristic rules, and its performance was ultimately evaluated through human assessment, achieving an accuracy of 92.75\%. Additionally, the model also identifies the sentiment polarity of the sentence's aspect. After processing the review text through a series of steps, we can analyze peer review texts from recent years at a fine-grained level, including word, sentence, and aspect-based perspectives. \\
\indent In the method proposed by Yuan et al. \cite{r8}, aspect identification is treated as a sequence labeling task. Specifically, given a review sentence consisting of $n$ words $S=w_1,...,w_n$, the objective is to convey appropriate aspect information in the input sequence through a mapping function. First, BERT is used to represent the sentence $S$ as a sequence containing $n$ tokens $(e_1,e_2,...,e_n)$, converting it into a context-enriched representation. This sequence is then fed into the trained mapping function to obtain feature classifications for labeling:\\
\begin{equation}
	p_i=softmax(We_i+b)
\end{equation}
Where $W$ and $b$ are trained parameters of the multilayer perceptron. $p_i$ is a vector that represents the probability of token $i$ being assigned to different aspects.\\
\indent Specifically, given a review text, we first split it into sentences and identify the aspect mentioned in each sentence. For all aspects except Summary, sentiment polarity (positive or negative) is also annotated. In Section \ref{sec4.1.3} and \ref{sec4.1.4}, for each year, we compute the average number of aspect mentions and the corresponding sentiment distributions across review texts.
For example, consider a review text containing 10 sentences, in which Summary is mentioned twice, and Originality and Clarity are each mentioned once, with Originality expressed positively and Clarity negatively. The corresponding dictionary used for aggregation can be represented as:
\{summary: 2, originality\_positive: 1, originality\_negative: 0, clarity\_positive: 0, clarity\_negative: 1, replicability\_positive: 0, replicability\_negative: 0, soundness\_positive: 0, soundness\_negative: 0, motivation\_positive: 0, motivation\_negative: 0, substance\_positive:0, substance\_negative: 0, meaningful\_comparison\_positive: 0, meaningful\_comparison\_negative: 0\}. Each review text is converted into such a dictionary, which is then used for subsequent analyses.

\subsection{LLM-Assisted Peer Review Text Detection}
To identify which review reports are LLM-assisted, we need to detect whether the review texts have been modified or generated by LLMs. In contrast to the previous studies, we adopted a more refined analytical approach, focusing on the characteristics at the word, sentence, and aspect levels, while Liang et al. \cite{r19} and Latona et al. \cite{r20} primarily conducted analysis at the overall text level. This methodological difference enables our research to provide a deeper understanding and more detailed analysis of the text. It is important to note that this study detects review comments that may have been assisted by LLMs, rather than those entirely generated by LLMs. We employ the maximum likelihood model designed by Liang et al. \cite{r19} to detect text that may have been assisted by artificial intelligence. We chose this model because our dataset has a very high similarity to the data used in the original study, ensuring that its performance remains valid for our case. This model leverages expert-authored and LLM-generated reference texts to accurately and efficiently assess corpus-level real-world usage of LLMs. And the average prediction error of the model is less than 5\%. The objective of this model is used maximum likelihood estimation (MLE)  $L(\alpha)$ to estimate the $\alpha$ score for LLM-generated content:
\begin{equation}
	L(\alpha) = \sum_{i=1}^{n}log((1-\alpha )P(x_i)+\alpha Q(x_i))
\end{equation}
Where $x$ to refer to a corpus, $P$ and $Q$ denote the probability distribution of documents written by scientists and generated by LLM in Liang et al. \cite{r19}'s model, respectively. We replace $x$ in $Q$ with our data to calculate the value of $\alpha$, thereby determining whether the input text is LLM-assisted generated. In addition to utilizing the model for detection, we also employed a lexicon of commonly and primarily used terms by LLMs, as developed in the studies by Liang et al. \cite{r19} and Latona et al. \cite{r20}, to aid in detection. The specific lexicon is provided in Appendix Table \ref{taba1}. Prior to applying the detection model described above, we first perform a filtering step using a predefined terminology dictionary. If the review text does not contain any terms from this dictionary, we assume it is unlikely to have been LLM-assisted and exclude it from further analysis. Only the texts that pass this initial filter are subsequently input into the detection model for evaluation.
\subsection{Lexical Sophistication and Syntactic Sophistication and Complexity for Review Text}
In addition to segmenting the review texts into word- and sentence-level units, we also calculated the lexical and syntactic complexity of the review texts for each year. Lexical complexity was measured using TAALES \cite{r46}, a tool that evaluates over 400 classic and novel indices of lexical complexity, including indices related to various substructures. Syntactic sophistication and complexity were assessed using the advanced syntactic analysis tool TAASSC \cite{r47}, which captures a wide range of metrics associated with syntactic development. These two tools do not calculate a single value to measure syntax but instead compute multiple distinct metrics for comparison. We selected several key metrics for our study, as shown in Table \ref{tab7} and \ref{tab8}. \\
\begin{table}[h]
	\vspace{-0.8cm}
	\caption{\centering{The lexical sophistication metrics used for peer review text.}}\label{tab7}%
	\centering
	\begin{tabular}{@{}m{45mm}<{\centering} m{40mm}<{\centering}m{40mm}<{\centering} @{}}
		\toprule
		\textbf{Metrics} & \textbf{Description}&\textbf{Types of Words} \\
		\midrule
		COCA Academic Range AW&	Mean Range (number of documents that a word occurs in) score&	All words\\
		COCA Academic Bigram Frequency&	Mean bigram frequency score&	All words\\
		COCA Academic Trigram Frequency&	Mean trigram frequency score&	All words\\
		
		\bottomrule 
	\end{tabular}
	\begin{tablenotes}
		\footnotesize
		\item \textbf{Note:} COCA is corpus of contemporary American English. AW is all words.
	\end{tablenotes}
\end{table}
\begin{table}[h]
	\vspace{-0.8cm}
	\caption{\centering{The syntactic sophistication and complexity metrics used for peer review. }}\label{tab8}%
	\centering
	\begin{tabular}{@{}m{45mm}<{\centering} m{40mm}<{\centering}m{40mm}<{\centering} @{}}
		\toprule
		\textbf{Metrics} & \textbf{Description}&\textbf{Type of Sentences} \\
		\midrule
		advcl\_per\_cl&	Number of adverbial clauses per clause&	Clause Complexity\\
		nsubj\_per\_cl&	Number of nominal subjects per clause&	Clause Complexity\\
		mark\_per\_cl&	Number of subordinating conjunctions per clause&	Clause Complexity\\
		aux\_per\_cl&	Number of auxilliary verbs per clause&	Clause Complexity\\
		dobj\_per\_cl&	Number of direct objects per clause&	Clause Complexity\\
		\bottomrule 
	\end{tabular}
\end{table}
\indent The reason for selecting these three metrics to measure lexical complexity is that they assess the frequency of a word's occurrence across different documents, reflecting the breadth of vocabulary usage. Additionally, analyzing the frequency of common phrases (bigrams and trigrams) reveals common expression structures and word collocations in the text. These metrics enable the evaluation of a text's lexical diversity, the richness of language expression, and sentence complexity. The five metrics chosen to calculate syntactic sophistication and complexity were selected because adverbial clauses, noun clauses as subjects, and subordinating conjunctions indicate sentence complexity and logical relationships. Auxiliary verbs, on the other hand, help to understand the reviewer's stance and tone, particularly in speculative or definitive judgments. Furthermore, the frequency of direct objects reflects the reviewer's attention to specific entities and the concreteness of the text. These metrics allow for a deeper analysis of the structural diversity, subjectivity, and rigor of argumentation in the review texts.
\section{Result}
In this section, we first present the trends in review text length and aspect mentions in ICLR and NeurIPS before and after the advent of LLMs. We then examine aspect evaluations by reviewers using LLM assistance versus those without, and finally, we analyze the relationship between LLM-assisted review texts and assigned scores.
\begin{figure*}[h]%
	\centering
	\includegraphics[width=1\textwidth]{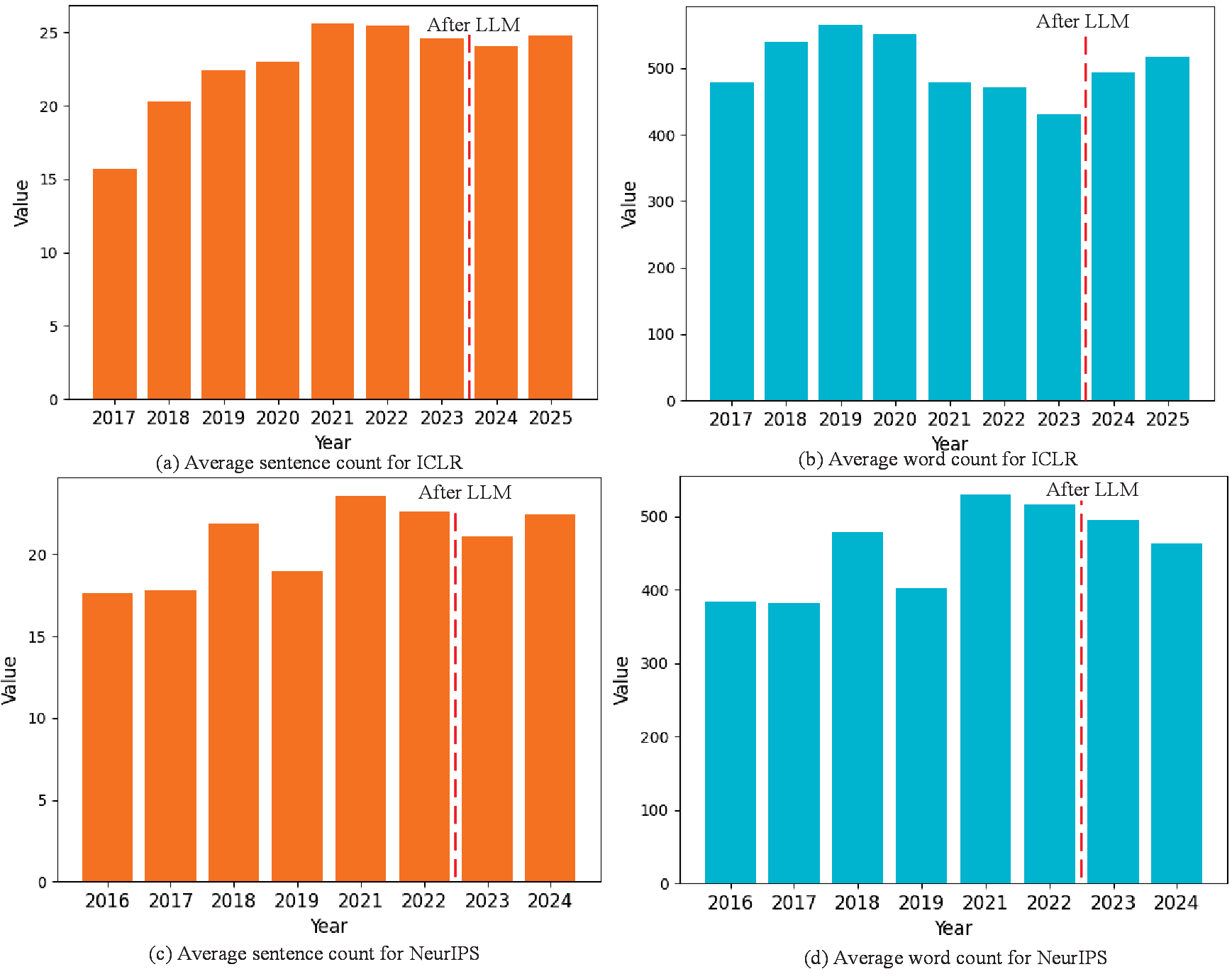}
	\caption{\centering{The trends in sentence and word lengths over the years for the ICLR and NeurIPS conferences.}}
	\label{fig:2}
\end{figure*}
\begin{figure*}[h]%
	\centering
	\includegraphics[width=1\textwidth]{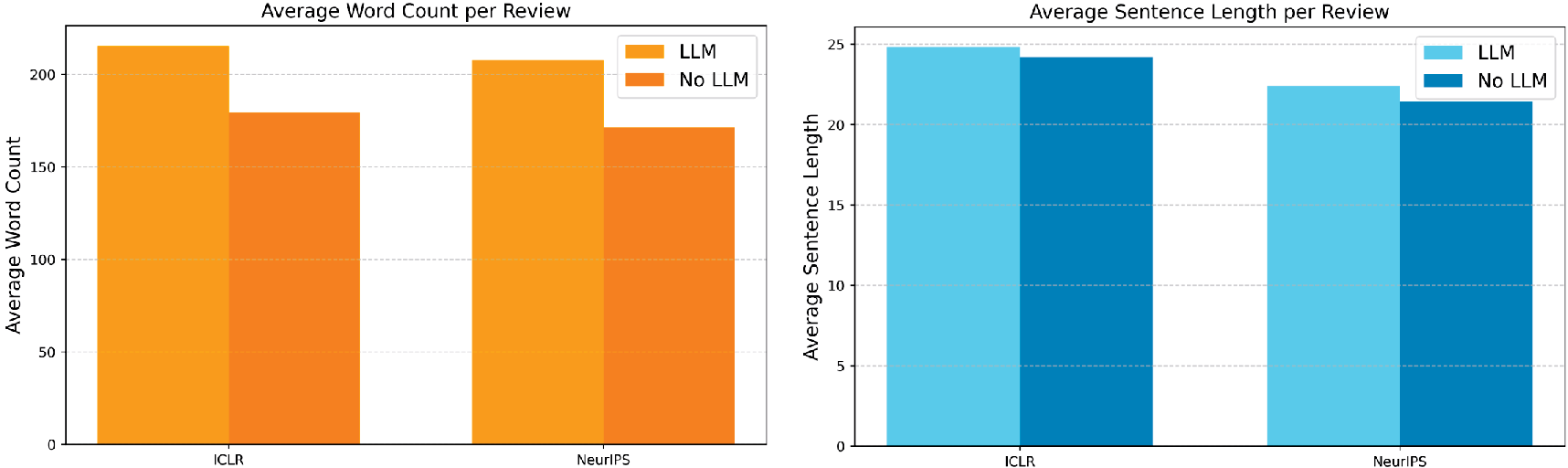}
	\caption{\centering{Average length of LLM assisted and non LLM assisted review texts in ICLR (2024-2025) and NeurIPS (2023-2024).}}
	\label{fig:ex}
\end{figure*}
\subsection{Impact of LLMs on Linguistic Complexity and Content-Level Expression}
This section addresses RQ1, which investigates how the emergence of LLMs has influenced the linguistic and content-level properties of peer review texts. The analysis examines both linguistic complexity (e.g., length, lexical and syntactic sophistication) and aspect-level expression (e.g., number, length, and sentiment of aspect mentions).

\subsubsection{Overall Text Length Patterns Before and After the Emergence of LLMs}

We first compare the length of reviews across pre- and post-LLM periods\footnote{The review release date for ICLR 2023 was November 5, 2022 (https://iclr.cc/Conferences/2023/Dates), and the rebuttal start date for NeurIPS 2022 was July 26, 2022 (https://nips.cc/Conferences/2022/Dates). In contrast, the rebuttal start date for NeurIPS 2023 was August 2, 2023 (https://nips.cc/Conferences/2023/Dates). Since ChatGPT was released on November 30, 2022, we consider the review processes of ICLR 2023 and NeurIPS 2022 to be unaffected by it, whereas reviews for subsequent ICLR and NeurIPS conferences may have been influenced by LLMs.}.
The trends in the length of review texts for ICLR and NeurIPS are illustrated in Figure \ref{fig:2}. The subgraph (a), (b), (c), and (d) in Figure \ref{fig:2} represent the average sentence count and average word count for ICLR, and the average sentence count and average word count for NeurIPS, respectively. The average sentence and word counts in ICLR reviews exhibit a rise-fall-rise pattern over time. This suggests that when LLMs first became available, only a small number of reviewers adopted them, resulting in a continuation of earlier trends. In contrast, ICLR 2025 \footnote{https://blog.iclr.cc/2025/04/15/leveraging-llm-feedback-to-enhance-review-quality/} explicitly mentioned the use of LLMs, indicating that reviewers assisted by LLMs were able to provide longer and more detailed feedback. Similarly, NeurIPS shows a comparable trend, but with noticeable differences in the average word count after the emergence of LLMs. A plausible explanation is that NeurIPS reviews more frequently correspond to accepted papers, which tend to include more concise yet targeted evaluations. To further examine the reasons behind the change in review length following the introduction of LLMs, we compared review texts from ICLR 2024-2025 and NeurIPS 2023-2024, distinguishing between LLM-assisted and non-assisted reviews, as illustrated in Figure \ref{fig:ex}. The results show that reviews potentially assisted by LLMs tend to be longer than those without LLM assistance, which aligns with the intuition that LLM-generated content typically increases text length. We think that the decline in review quality may be a contributing factor to the observed reduction in review length. Recent studies \cite{r42,r43,r44,r45} have indicated a gradual decrease in the quality of peer reviews over the years, and the trend in text length may serve as a tangible reflection of this decline in quality.

\subsubsection{Linguistic Complexity}
To capture the fine-grained changes in language use, we analyze lexical and syntactic complexity. Lexical richness reflects vocabulary diversity and sophistication, while syntactic indicators measure grammatical elaboration.
\begin{figure}[H]%
	\centering
	\includegraphics[width=1\textwidth]{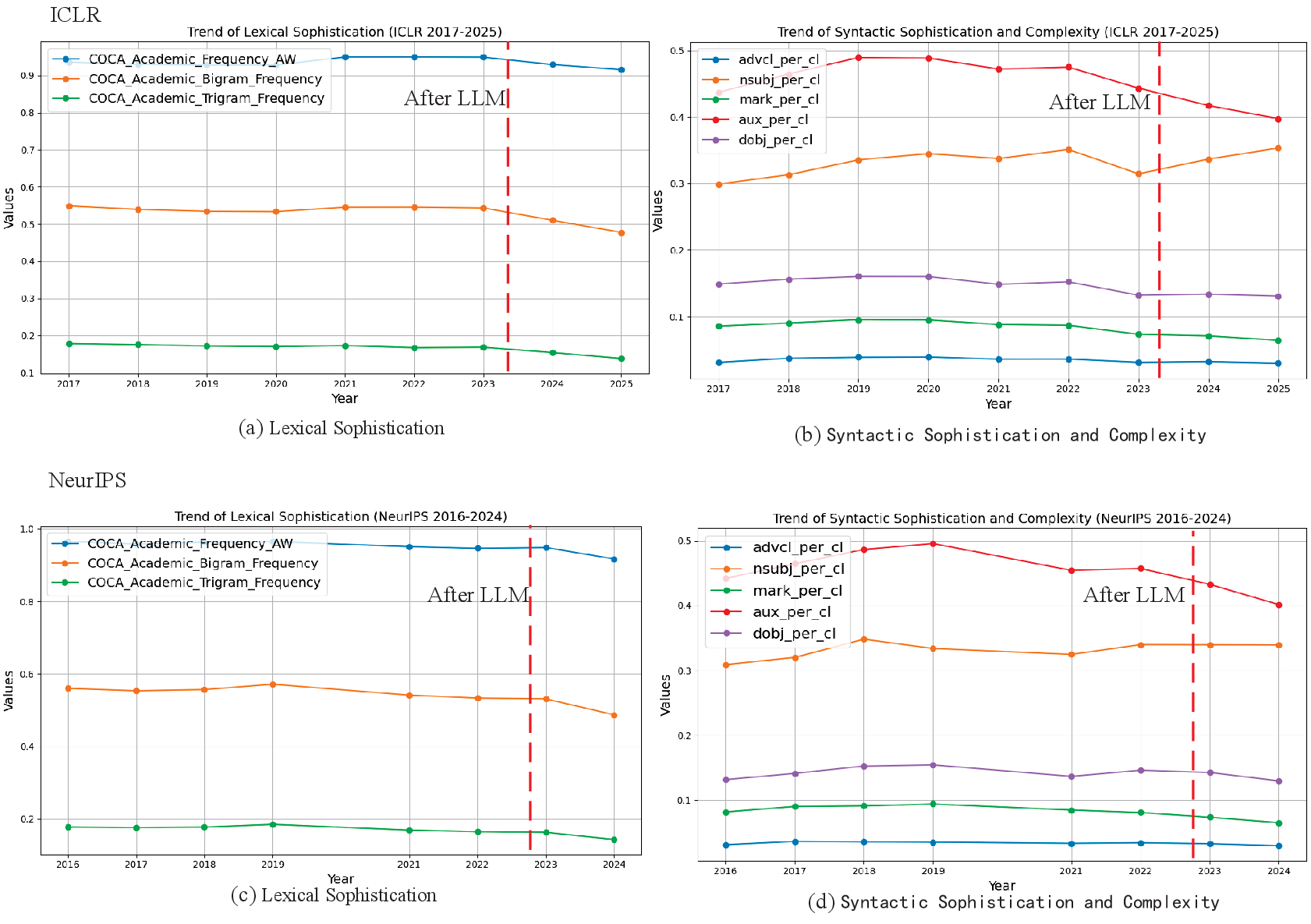}
	\caption{\centering{Lexical sophistication and syntactic sophistication and complexity of review texts in ICLR and NeurIPS.}}
	\label{fig:8}
\end{figure}
As shown in Figure \ref{fig:8}, the lexical complexity of reviews in both conferences remains relatively stable over time, though it exhibits a slight decline following the emergence of LLMs. In terms of syntactic complexity, both conferences display similar overall trends; however, a notable exception is the significant increase in the number of nominal subjects in ICLR reviews after the introduction of LLMs. We hypothesize that the decrease in lexical complexity observed in ICLR 2024 may be associated with the adoption of LLMs. LLMs are capable of generating concise, clear, and readable text while generally avoiding overly complex terminology, which may contribute to a reduction in lexical complexity within review reports. Recent studies \cite{r36,r48} have examined the applications and impacts of LLMs in academic writing. Their findings indicate that since the release of ChatGPT, the use of ChatGPT or other LLMs in abstracts has been steadily increasing, and by February 2024, approximately 35\% of arXiv abstracts may have been modified using ChatGPT or other LLMs. Peer review reports from ICLR 2024 and NeurIPS 2023 have also been influenced, to varying degrees, by the use of language learning models \cite{r19,r20}. The increase in nominal subject usage within ICLR reviews suggests that reviewers are becoming more inclined to employ noun phrases as subjects rather than personal pronouns (e.g., "I" or "we"), thereby enhancing the objectivity and professionalism of their writing. Additionally, we observe a downward trend in the use of auxiliary verbs across both conferences. A possible explanation for this trend is that LLMs tend to generate concise and straightforward sentences, leading to a lower frequency of auxiliary verb usage. Consequently, reviewers assisted by language learning models may have contributed to this overall decline.\\
\subsubsection{Aspect-Level Expression Patterns}
\label{sec4.1.3}
\begin{figure}[H]%
	\centering
	\includegraphics[width=1\textwidth]{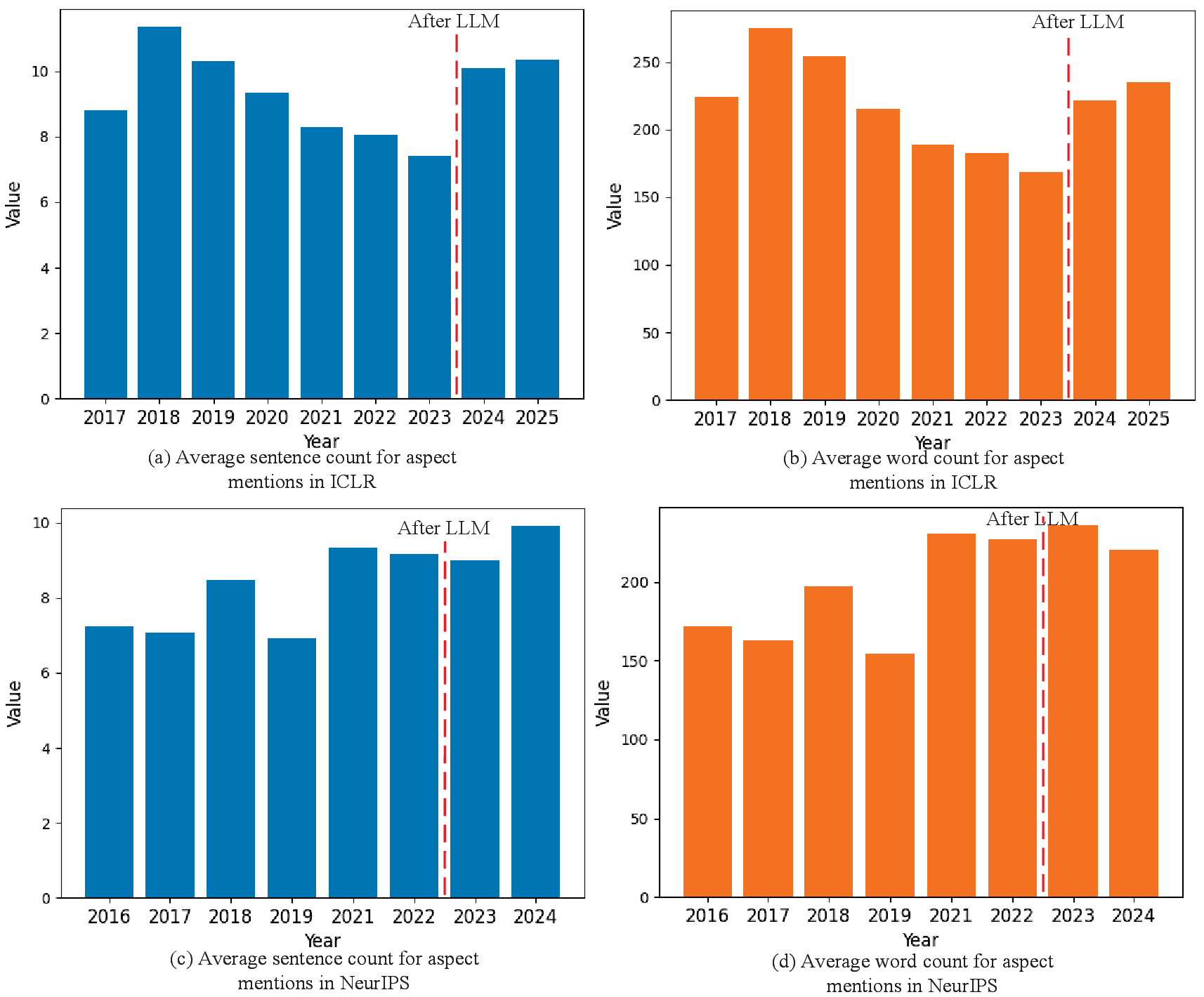}
	\caption{\centering{The average length of sentences and words related to the review aspects in ICLR 2017-2025 and NeurIPS 2016-2024 (except 2020).}}
	\label{fig:4}
\end{figure}

\begin{figure}[H]%
	\centering
	\includegraphics[width=0.75\textwidth]{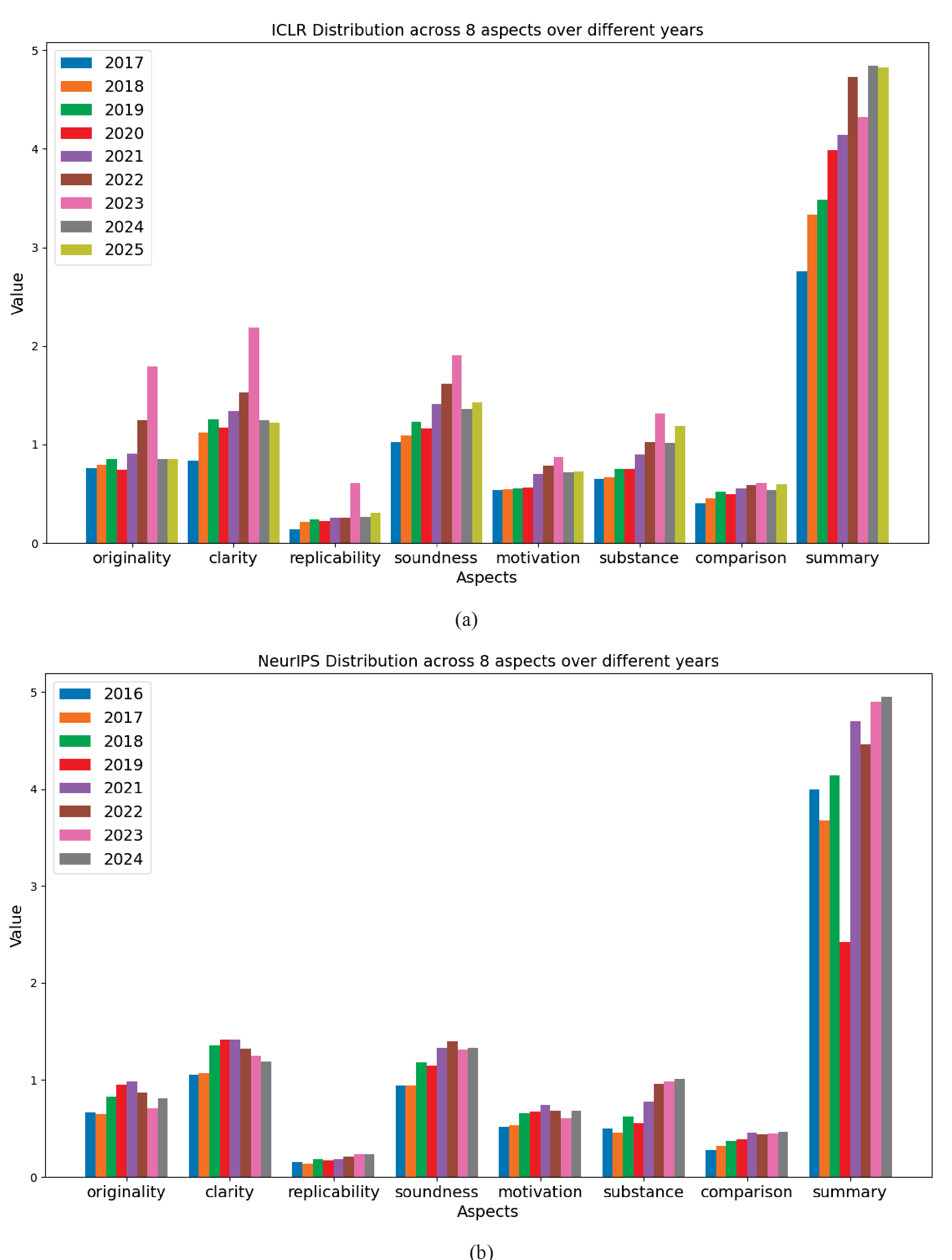}
	\caption{\centering{The distribution of aspect mentions and their sentiment in ICLR 2017-2025 and NeurIPS 2016-2024. Comparison indicates a meaningful comparison.}}
	\label{fig:asp}
\end{figure}

\begin{figure}[H]%
	\centering
	\includegraphics[width=0.95\textwidth]{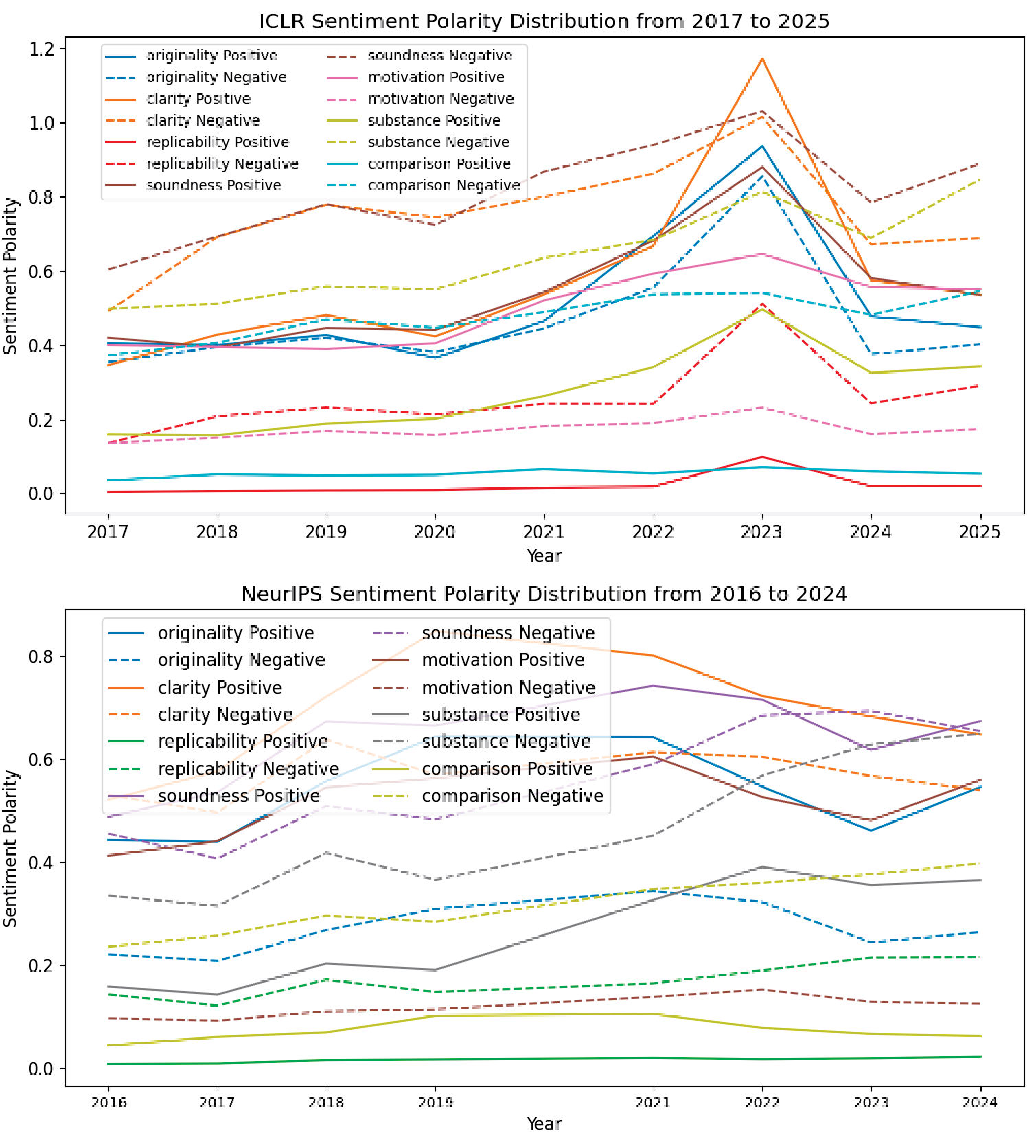}
	\caption{\centering{The sentiment distribution of aspect mentions in ICLR 2017-2025 and NeurIPS 2016-2024 (except 2020). Comparison indicates a meaningful comparison.}}
	\label{fig:sentiment}
\end{figure}

We then examine how content expression has shifted at the aspect level. Specifically, we analyze: The average length of aspect-related contents to measure elaboration depth (Figure \ref{fig:4}). The average number of aspect mentions (e.g., Clarity, Soundness, Originality) to assess topic coverage (Figure \ref{fig:asp}).  The average sentiment polarity associated with each aspect to explore tonal tendencies (Figure \ref{fig:sentiment}). The detailed computation procedures are provided in Section \ref{sec3.2}. \\

As shown in Figure \ref{fig:4}, after 2018, the average sentence and word lengths of aspect-related mentions in ICLR reviews exhibit an overall declining trend, followed by a noticeable increase after the emergence of LLMs. In contrast, the variation in NeurIPS is less pronounced, but a similar upward trend in sentence count can be observed after the emergence of LLMs. This suggests that the advent of LLMs may have influenced the way reviewers compose their reports and may even have encouraged some reviewers to employ LLMs as assistance in conducting aspect-level evaluations of papers. In addition, the results in the figure reveal a sharp increase from 2017 to 2018. We attribute this phenomenon to the emergence of the Attention mechanism \cite{r59} in 2017, which triggered a rapid expansion of research activity in the deep learning community, thereby leading to the observed surge during this period.
Then, we analyzed the overall aspect mentions in ICLR and NeurIPS peer review texts before and after the advent of LLMs. As shown in Figure \ref{fig:asp}, we present the annual distribution of reviewers' focus across various aspects. Overall, both venues consistently emphasize summary, which remains the dominant linguistic component across years. The recent rise in the frequency of summary mentions may be attributed to the growing complexity of academic papers, prompting reviewers to produce longer and more detailed summaries. For ICLR, a pronounced peak is observed in clarity, soundness, and summary between 2022 and 2023. This sharp increase coincides with the early diffusion of LLM-assisted reviewing, suggesting that automated or semi-automated tools enhanced linguistic fluency and logical coherence. However, after 2023, both conferences show a decline in the frequency of originality-related evaluations, which we hypothesize may also be influenced by the use of LLMs. In contrast, NeurIPS exhibits a relatively stable pattern with only slight increases, indicating a more standardized and less volatile reviewing style. A possible explanation for this discrepancy is that, in our dataset, the number of rejected papers from ICLR is substantially higher than that from NeurIPS. Overall, the results suggest that the emergence of LLMs has had a positive impact on the clarity and soundness dimensions of reviews, yet it has not substantially improved meaningful comparison or replicability. This finding implies that automated tools primarily enhance expressive fluency rather than reviewers' critical reasoning capabilities.

Finally, the sentiment polarity distribution for the above aspects is illustrated in Figure \ref{fig:sentiment}. A notable shift in sentiment can be observed in ICLR between 2022 and 2024. This suggests that as submission volumes surged, reviewers exhibited changes in emotional tone. With the emergence of LLMs, while submission volumes continued to rise, reviewers gained access to tools potentially enhancing their efficiency, likely contributing to increased emotional stability. The sentiment distribution for NeurIPS demonstrates relatively stable trends, which aligns with the steady aspect mention patterns.

\subsubsection{Interaction Between Reviewer Confidence and Linguistic Features}
\label{sec4.1.4}
\begin{figure}[H]%
	\centering
	\includegraphics[width=1\textwidth]{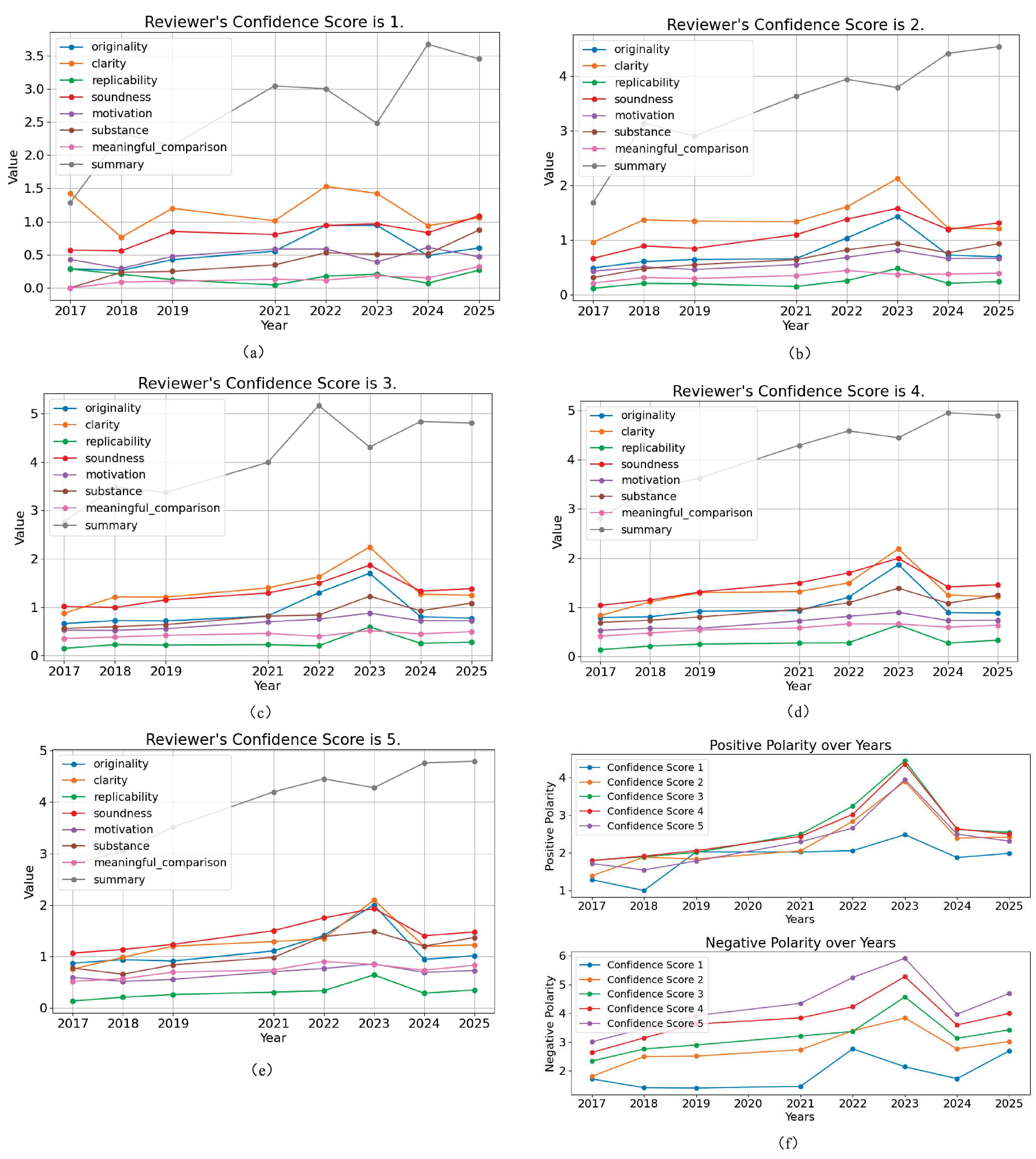}
	\caption{\centering{Distribution of aspect mentions and sentiment polarity across different confidence scores in ICLR (2017-2025, except 2020). Figures (a)-(e) represent the distribution of aspect mentions for confidence scores ranging from 1 to 5, while figure (f) shows the sentiment polarity distribution.}}
	\label{fig:6}
\end{figure}

\begin{figure}[H]%
	\centering
	\includegraphics[width=1\textwidth]{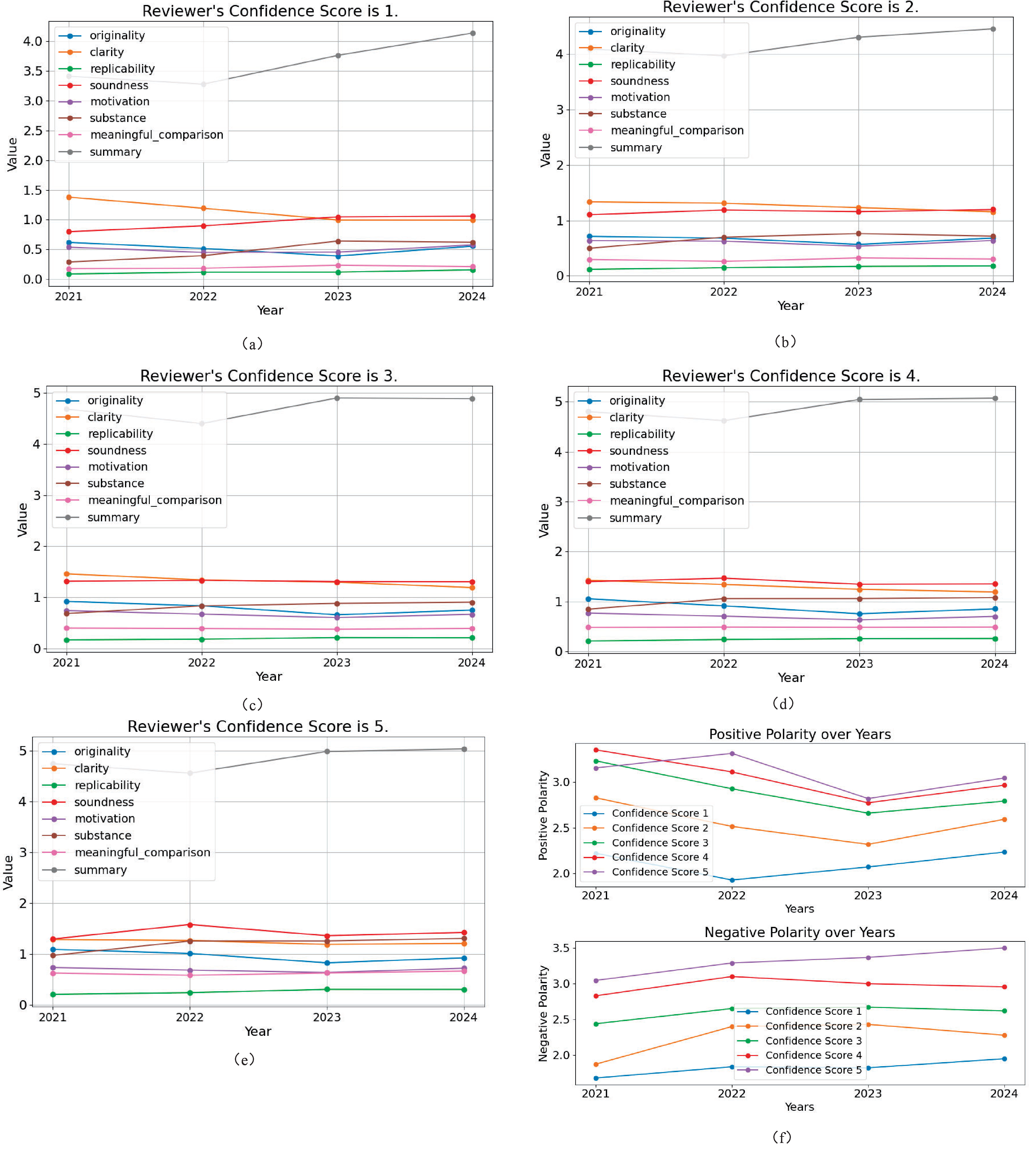}
	\caption{\centering{Distribution of aspect mentions and sentiment polarity across different confidence cores in NeurIPS (2021-2024). Figures (a)-(e) represent the distribution of aspect mentions for confidence scores ranging from 1 to 5, while figure (f) shows the sentiment polarity distribution.}}
	\label{fig:7}
\end{figure}

Then, we test whether the above linguistic and content-level changes vary by reviewers' self-reported confidence scores. This analysis reveals whether LLMs amplify or mitigate stylistic and expressive differences between confident and less-confident reviewers.
\begin{figure}[H]%
	\centering
	\includegraphics[width=1\textwidth]{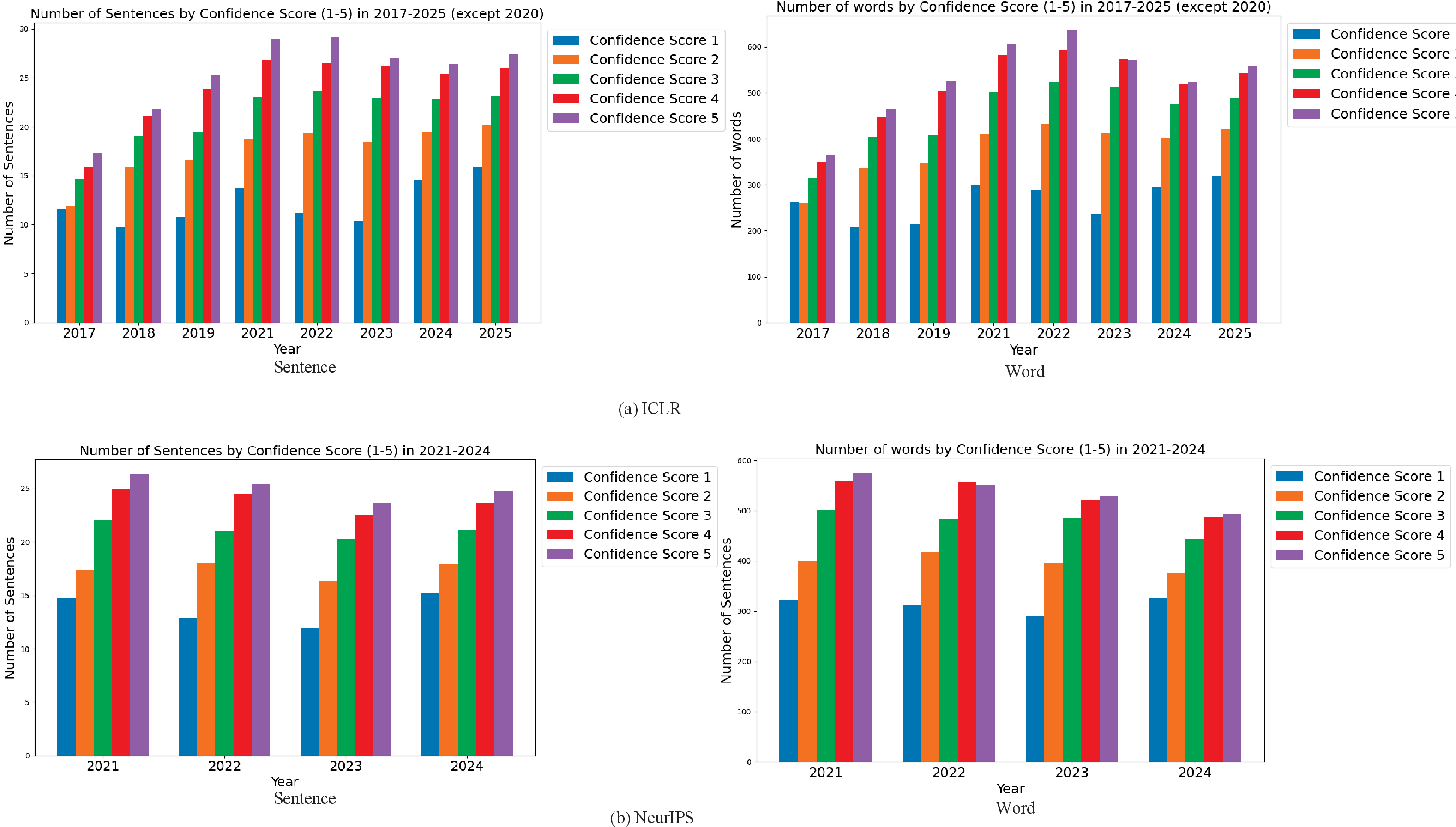}
	\caption{\centering{Number of words and sentences by confidence scores (1-5) in ICLR (2017-2025, except 2020) and NeurIPS (2021-2024).}}
	\label{fig:5}
\end{figure}
Firstly, we analyze the average length variation of review reports with different confidence scores. As shown in Figure \ref{fig:5}(a), we observe that ICLR reviewers with confidence scores between 2 and 5 exhibit a trend where average review length initially increases and then decreases. Meanwhile, reviewers with a confidence score of 1 show an upward trend after 2023, indicating a differing trend from other confidence scores. In Figure \ref{fig:5}(b), we see that NeurIPS reviewers with varying confidence scores generally follow similar trends, except for those with a confidence score of 1, whose patterns resemble those observed for ICLR. From these observations, we infer that while higher-confidence reviewers tend to write longer reports, the length of these reports has been decreasing in recent years. This indicates that high confidence score reviewers become increasingly familiar with the field over time, especially after the emergence of LLMs. High confidence score reviewers can write more accurate and concise evaluations that directly address key issues, resulting in shorter overall lengths. The observed upward trend among reviewers with a confidence score of 1 indicates that the emergence of LLMs can help reviewers with low confidence scores learn more about areas and knowledge they are not familiar with. This enables them to provide relatively informed reviews instead of submitting low value evaluations due to unfamiliarity with the topic, as in the past.
\begin{figure}[H]%
	\centering
	\includegraphics[width=1\textwidth]{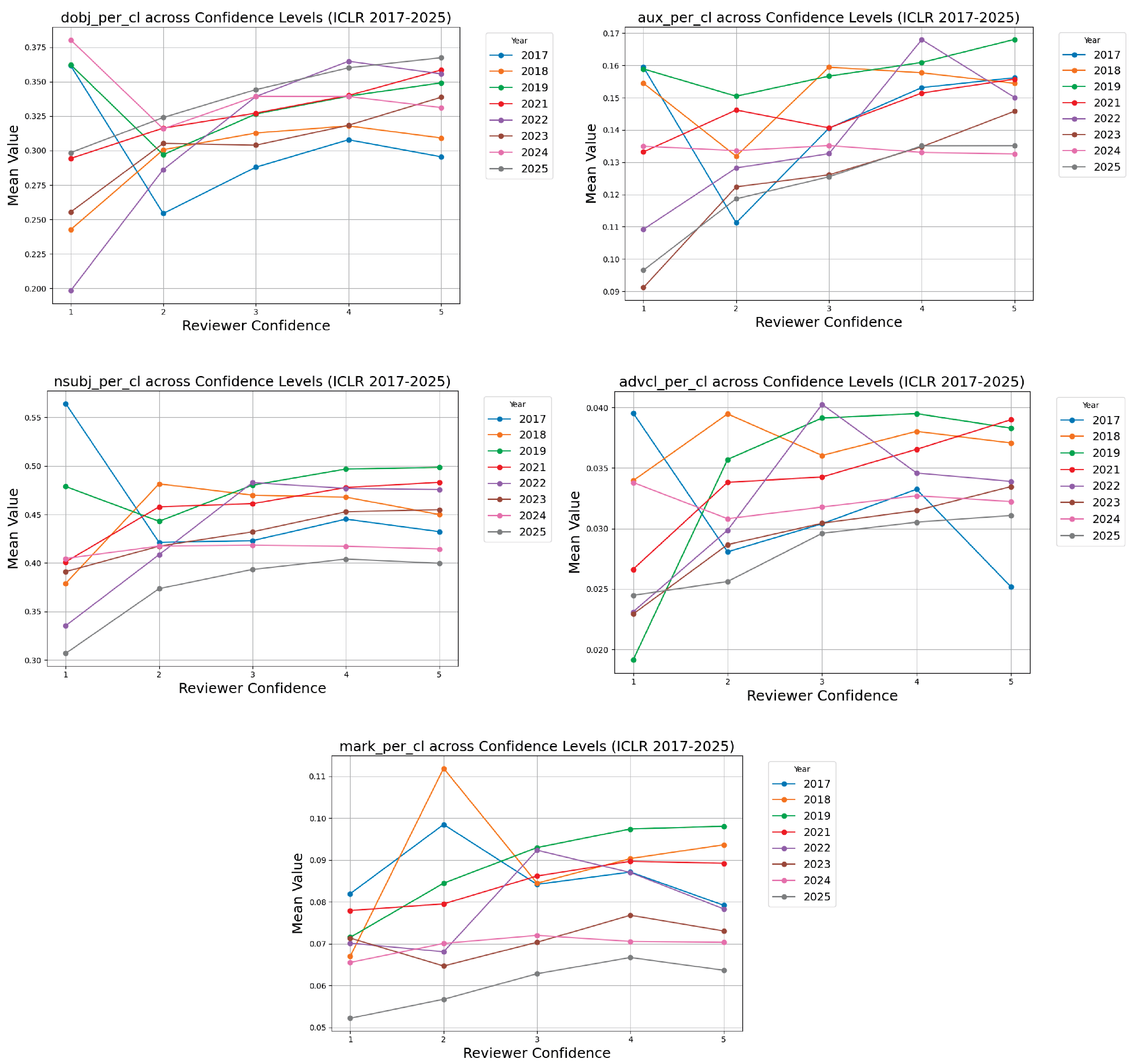}
	\caption{\centering{Yearly trends of syntactic sophistication and complexity across reviewer confidence levels in ICLR (2017-2025, except 2020).}}
	\label{fig:sen_iclr}
\end{figure}
\begin{figure}[H]%
	\centering
	\includegraphics[width=1\textwidth]{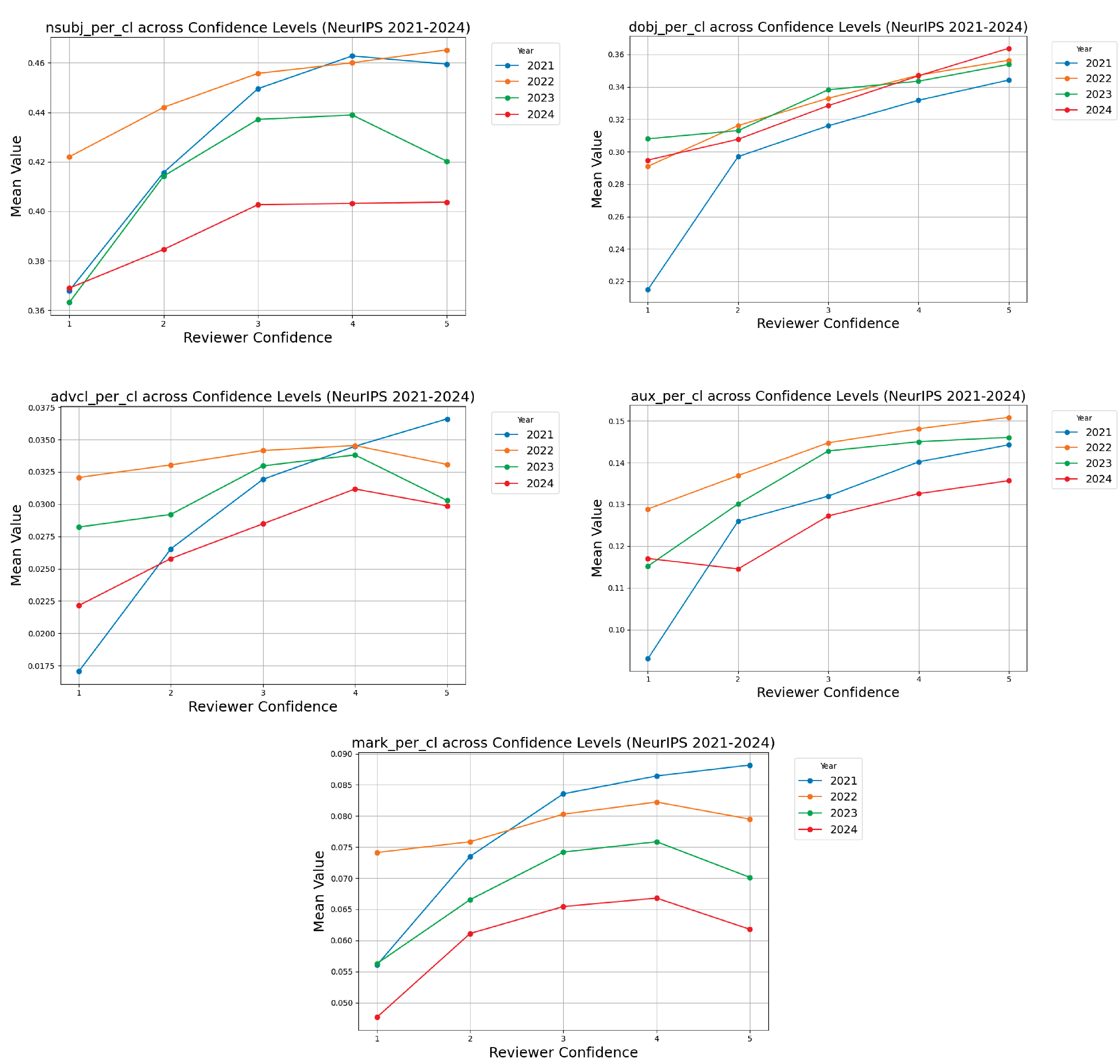}
	\caption{\centering{Yearly trends of syntactic sophistication and complexity across reviewer confidence levels in NeurIPS (2021-2024).}}
	\label{fig:sen_nips}
\end{figure}
We then analyze the lexical and syntactic complexity of review texts with different levels of reviewer confidence. Figure \ref{fig:sen_iclr} and \ref{fig:sen_nips} present the variation of syntactic sophistication and complexity across reviewer confidence levels for ICLR (2017-2025, except 2020) and NeurIPS (2021-2024). Overall, reviews with higher confidence tend to exhibit greater syntactic complexity, characterized by a higher proportion of auxiliary verbs (aux\_per\_cl) and adverbial clauses (advcl\_per\_cl), reflecting more elaborated and logically structured sentences. In contrast, features such as direct objects (dobj\_per\_cl) and clause markers (mark\_per\_cl) show more fluctuation, indicating stylistic diversity among reviewers. After the emergence of LLMs (notably ICLR 2024-2025 and NeurIPS 2024), the syntactic patterns become more stable and consistent across confidence levels. Specifically, the increased use of auxiliaries and adverbial clauses suggests that LLM-assisted reviews employ more standardized and cohesive syntactic constructions, while the slight decline in direct object and marker frequencies implies reduced syntactic depth and less structural variety. Compared with NeurIPS, ICLR demonstrates a more pronounced increase in syntactic regularity, highlighting a stronger influence of LLM assistance on the linguistic formulation of review texts. These findings indicate that LLMs may contribute to grammatically refined yet more homogeneous sentence structures, reinforcing fluency but potentially diminishing individual variation in writing style.
\begin{figure}[H]%
	\centering
	\includegraphics[width=1\textwidth]{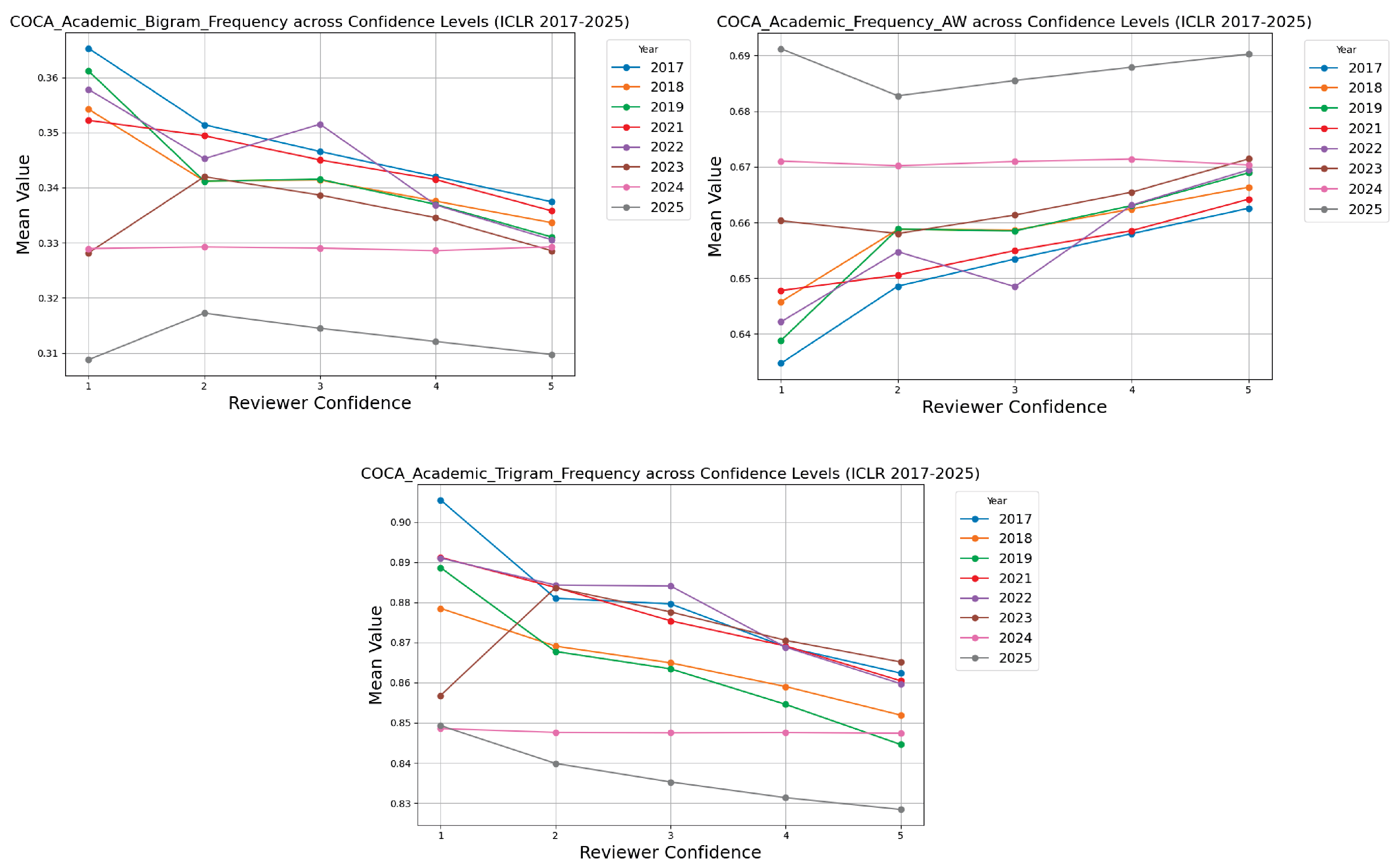}
	\caption{\centering{Yearly trends of lexical sophistication across reviewer confidence levels in ICLR (2017-2025, except 2020).}}
	\label{fig:lex_iclr}
\end{figure}
\begin{figure}[H]%
	\centering
	\includegraphics[width=1\textwidth]{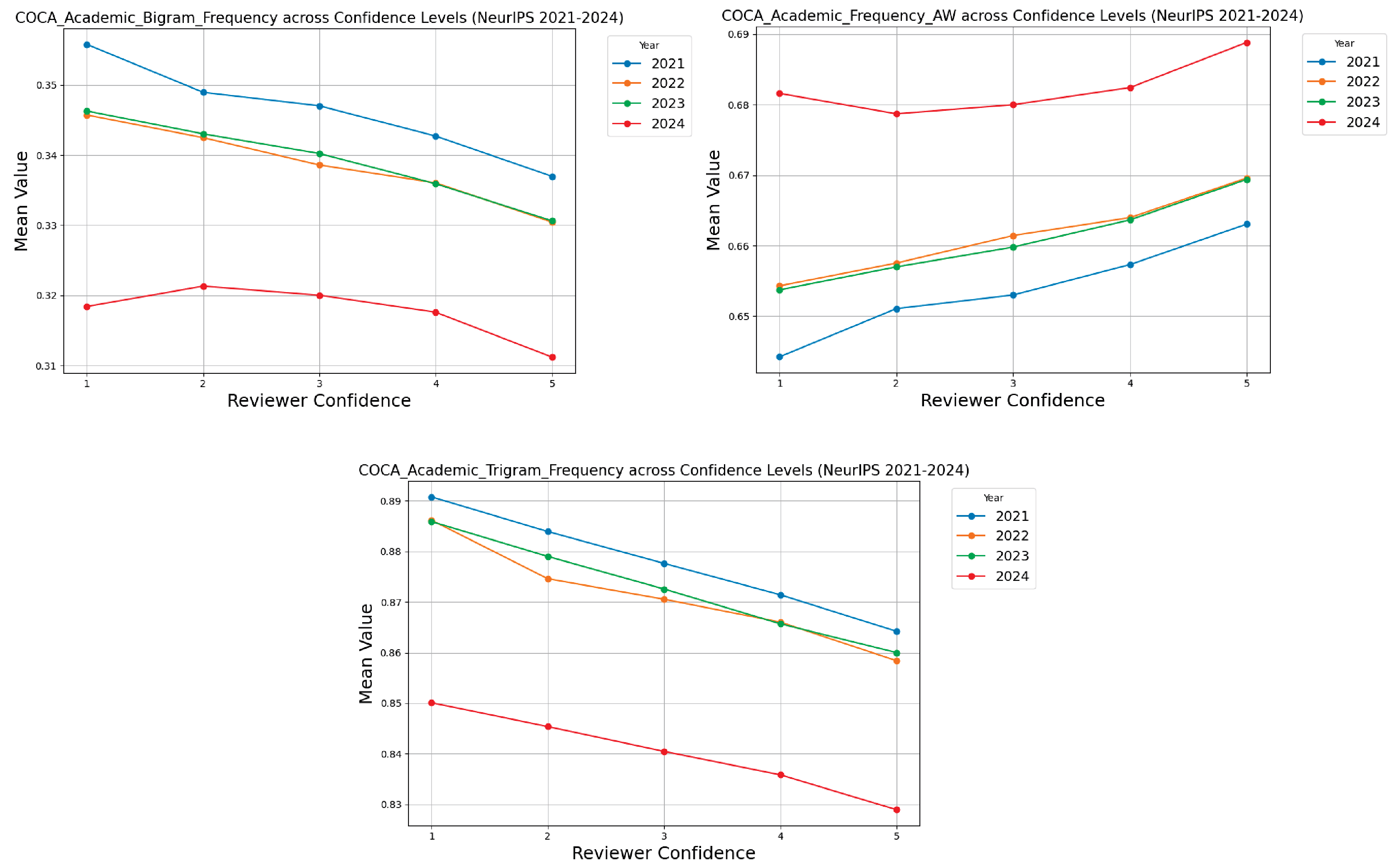}
	\caption{\centering{Yearly trends of lexical sophistication across reviewer confidence levels in NeurIPS (2021-2024).}}
	\label{fig:lex_nips}
\end{figure}
Figure \ref{fig:lex_iclr} and \ref{fig:lex_nips} present the variation in academic lexical frequencies across reviewer confidence levels for ICLR (2017-2025, except 2020) and NeurIPS (2021-2024). In both venues, reviews with higher confidence generally show lower bigram and trigram frequencies, indicating that more confident reviewers tend to employ less formulaic and more flexible academic language. However, after the introduction of LLMs (notably in ICLR 2024-2025 and NeurIPS 2024), these differences become attenuated: academic word frequency increases while bigram/trigram frequency declines, suggesting that LLM-assisted reviews adopt more standardized yet less lexically diverse expressions. Compared with NeurIPS, ICLR demonstrates a clearer rise in academic word frequency in the LLM period, implying a stronger influence of model-assisted writing on linguistic normalization.\\
\indent Finally, we analyzed the aspect mentions and sentiment polarity distribution for reviewers with different confidence scores, as shown in Figures \ref{fig:6} and \ref{fig:7}. For these two figures, we first briefly describe the criteria used for grouping by year and aspect. For example, Figure \ref{fig:6}(a) shows the number of review texts from ICLR 2021 with a confidence score of 1 (the number of reviews with a confidence score of 1 in that year is 92, as reported in Table \ref{tab:taba3}). We identify the number of mentions of each aspect in every review text and the corresponding sentiment polarity, and compute their averages; accordingly, each value shown for an aspect in the figure represents the average number of mentions. In addition, for sentiment polarity (Figure \ref{fig:6}(f)), we sum the sentiment polarity of all aspects within each review text for that year (positive and negative are calculated separately), and then compute the average sentiment across all review texts with different confidence scores in that year.\\

\indent Firstly, in Figures \ref{fig:6} (a)-(e), we present the average number of aspects mentioned by reviewers with confidence scores ranging from 1 to 5 in ICLR. The results indicate that, aside from the aspect of summary, the trends in mentions of other aspects correlate with the changes in average length. Additionally, we observe that regardless of the confidence score, mentions of the summary have shown an upward trend after 2023, particularly pronounced among reviewers with a confidence score of 1. We think that reviewers with a confidence score of 1 may be less familiar with the review domain, thus relying more on the abstract section of the paper for their evaluation. Furthermore, after 2023, LLM-assisted review may have placed greater emphasis on the abstract, or the generated review reports may have highlighted the summary more prominently, facilitating easier reference for reviewers, particularly those who are less knowledgeable about the field. This also suggests that reviewers utilizing LLM assistance may primarily focus on the evaluation of the summary rather than assessing core aspects of the paper, such as originality. We also present the average number of aspects mentioned by reviewers with confidence scores ranging from 1 to 5 in NeurIPS in Figures \ref{fig:7}(a)-(e). From the results in the figure, we can observe an upward trend in the number of mentions of the summary by reviewers across all confidence scores in 2023. However, the average length in NeurIPS that year was able to accommodate this increasing trend. Therefore, we infer that NeurIPS reviewers may have been less affected by LLM assistance.\\
\indent We also present the sentiment polarity changes for reviewers with different confidence scores at ICLR and NeurIPS, as shown in Figures \ref{fig:6}(f) and \ref{fig:7}(f). From the figures, we observe that reviewers with higher confidence scores (scores of 4 and 5) exhibit more pronounced sentiment fluctuations, particularly in negative sentiment. LLM may assist these high-confidence reviewers in conducting more detailed and in-depth assessments, enabling them to identify more issues within the papers, thereby increasing the expression of negative sentiment. These reviewers may use LLM for efficient critical evaluation. On the other hand, reviewers with lower confidence scores (scores of 1 and 2) display significant sentiment variation, which could reflect their useness on LLM during the review process. Given their potential lack of deep understanding in the field, LLM may partially compensate for their knowledge gaps, assisting them in analyzing the content of the papers and resulting in noticeable shifts in sentiment polarity. The above analysis assumes that reviewers are utilizing LLM-assisted tools during the review process. Previous studies by Liang et al. \cite{r19} and Latona et al. \cite{r20} have indicated that at least 15\% of review reports were LLM-assisted.\\
\begin{figure}[H]%
	\centering
	\includegraphics[width=1\textwidth]{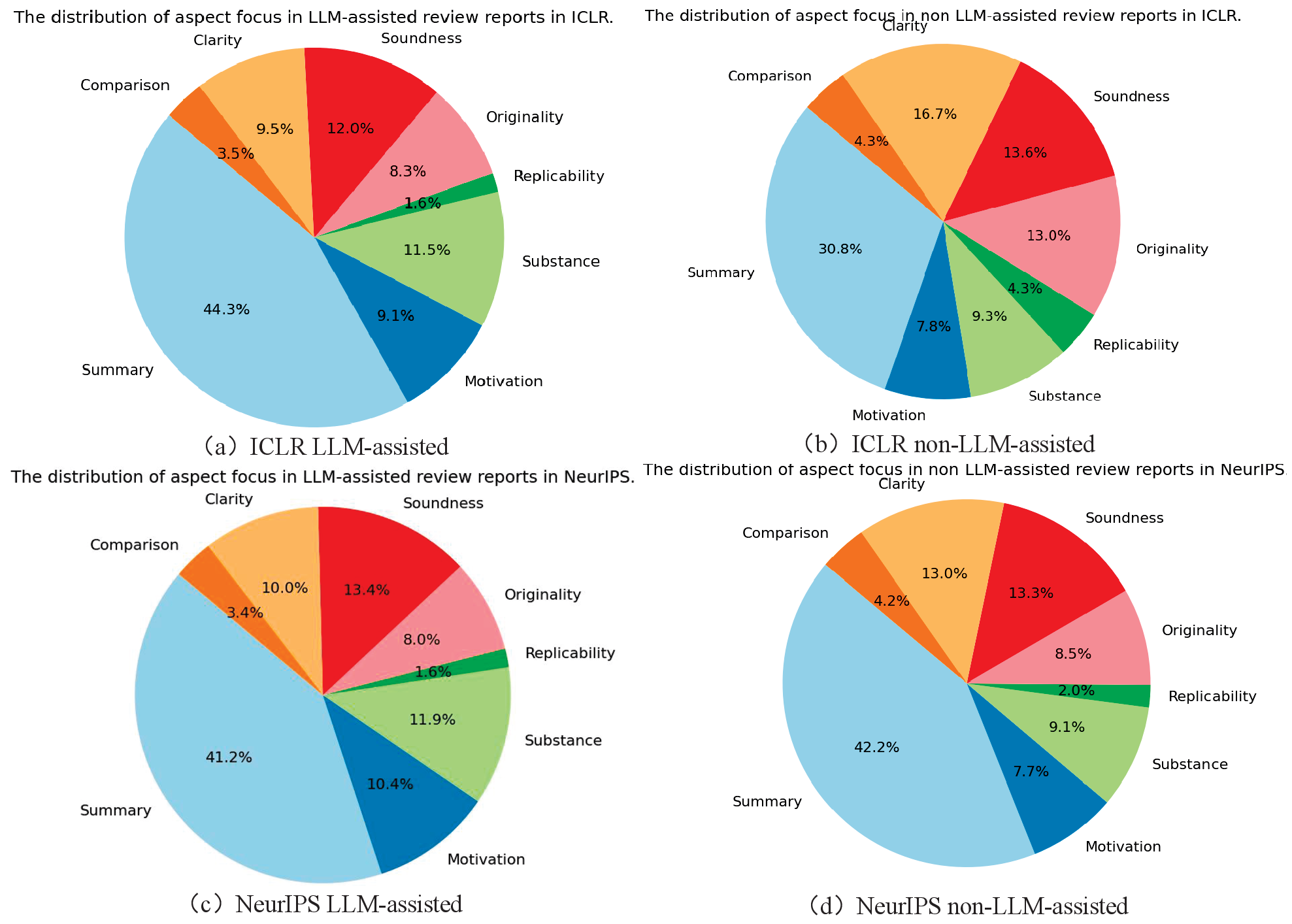}
	\caption{\centering{The distribution of aspect mentions between LLM-assisted reviewers and non-LLM-assisted reviewers. Comparison indicates a meaningful comparison.}}
	\label{fig:9}
\end{figure}
\subsection{Which Evaluation Aspects Are More Prominently Reflected in LLM-assisted Reviews}
To address RQ2, we conducted LLM-assistance detection on the review reports from ICLR 2024-2025 and NeurIPS 2023-2024, obtaining 9775 and 5884 samples, respectively. We then performed aspect recognition on the reports identified as LLM-assisted. Similarly, we carried out aspect recognition on an equivalent number of reports that were not identified as LLM-assisted, and compared the aspect mention distributions between the two, as shown in Figure \ref{fig:9}. The non-LLM-assisted peer review texts were sampled from ICLR 2023 and NeurIPS 2022, with the number of samples matched to those of ICLR 2024-2025 and NeurIPS 2023-2024, respectively.\\
\indent As shown in Figure \ref{fig:9}, in both ICLR and NeurIPS, the summary section occupies the largest proportion of content in both LLM-assisted and non-LLM-assisted review reports, highlighting its central role in reviewers' evaluation processes. However, in ICLR reviews, LLM-assisted reports tend to allocate a larger share to the summary compared to non-LLM-assisted ones, whereas the opposite pattern is observed in NeurIPS, although the difference there is less pronounced than in ICLR. We attribute this discrepancy to differences in data composition: in our sample, the vast majority of NeurIPS reviews correspond to accepted papers, while ICLR includes a substantial number of rejected submissions, which may lead to different distributions of emphasis on summaries. Overall, these results indicate that LLM assistance does not systematically increase reviewers' emphasis on the summary section. The motivation aspect also receives slightly greater emphasis in LLM-assisted reviews, indicating that reviewers may rely on generative models to elaborate on the contextual significance of a study.\\
\indent A similar pattern is observed for NeurIPS, where both substance and originality receive comparable levels of attention across review types, though originality tends to be marginally higher in non-LLM-assisted reports. Conversely, LLM-assisted reports generally devote less focus to clarity, particularly in ICLR, implying that the linguistic fluency provided by LLMs may reduce reviewers' perceived need to explicitly comment on the readability or organization of the paper. Furthermore, non-LLM-assisted reports place relatively greater emphasis on replicability and soundness than LLM-assisted ones, suggesting that human reviewers continue to engage more with methodological robustness and reproducibility concerns dimensions that are less frequently discussed in LLM-assisted writing.\\
\indent Notably, while NeurIPS exhibits limited distributional change across LLM-assisted and non-LLM-assisted groups, ICLR demonstrates more pronounced shifts. Specifically, LLM-assisted reviewers in ICLR show heightened attention to summary, substance, and replicability, while their mentions of originality and clarity decrease considerably. Non-LLM-assisted reviewers, by contrast, focus primarily on summary, clarity, soundness, and originality, whereas LLM-assisted reviewers emphasize summary, soundness, substance, and clarity.\\
\indent This shift suggests that, with the aid of LLM, reviewers tend to generate broader and more structured overviews, potentially improving their comprehension of a paper's content. However, such assistance appears to come at the cost of reduced engagement with critical and creative dimensions of evaluation, such as originality and replicability. The declining attention to originality, in particular, raises concerns about whether LLM-assisted reviews may inadvertently promote uniformity and reduce the diversity of evaluative perspectives within peer review discourse.

\subsection{Effects of LLM Assistance on Reviewers' Scoring and Confidence}
To address RQ3, we calculated the correlation between aspect mentions by LLM-assisted reviewers and the scores they assigned, including their confidence scores, using the Spearman correlation coefficient.
\subsubsection{Correlation Analysis Between Aspect Mentions and Reviewer-Assigned Scores}
First, we investigate the relationship between the aspects mentioned in review texts and the scores assigned by reviewers. Specifically, we examine whether the frequency with which different review aspects are discussed is associated with reviewers' overall scores. To this end, we conduct a correlation analysis and provide visualizations in the form of scatter plots to illustrate the distributional patterns between aspect mentions and reviewer-assigned scores across ICLR and NeurIPS.
\begin{figure}[H]%
	\centering
	\includegraphics[width=1\textwidth]{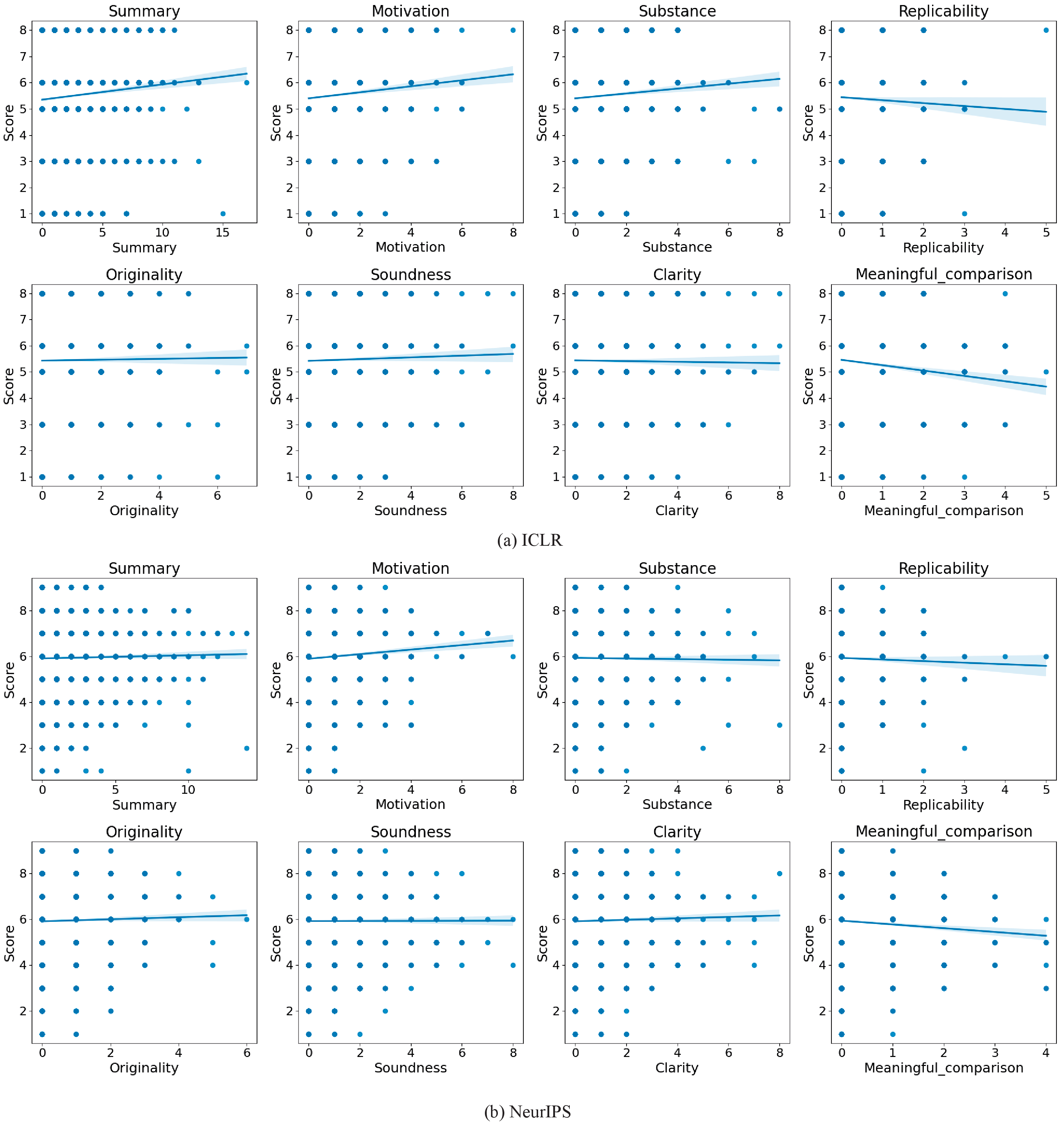}
	\caption{\centering{Scatter plots of overall score versus aspects mentioned in ICLR and NeurIPS.}
	\textbf{Note:} The solid lines represent simple linear regression fits and the shaded areas show the associated confidence intervals. The figure is intended as a visual aid for qualitative comparison rather than as the basis for formal regression analysis.}
	\label{fig:10}

\end{figure}
\begin{landscape}
	
	\begin{table}[t]
		\centering
		\caption{The results of correlation analysis between aspect mentions and overall scores in ICLR and NeurIPS. Summ represents Summary, Moti represents Motivation, Rep represents Replicability, Ori represents Originality, Meaningful represents Meaningful comparison.}
		\label{tab9}
		\large
		\begin{tabular}{@{}m{10mm}<{\centering} m{15mm}<{\centering} m{18mm}<{\centering} m{18mm}<{\centering} m{18mm}<{\centering} m{20mm}<{\centering} m{18mm}<{\centering} m{18mm}<{\centering} m{18mm}<{\centering} m{20mm}<{\centering}@{}}
			\toprule
			\textbf{Venue} & & \textbf{Summ} & \textbf{Moti} & \textbf{Substance} & \textbf{Rep} & \textbf{Ori} & \textbf{Soundness} & \textbf{Clarity} & \textbf{Meaningful} \\
			\midrule
			ICLR & score & 0.04*** & 0.05*** & 0.04*** & -0.02** & 0.01 & 0.01 & 0.00 & -0.05*** \\
			NeurIPS & score & 0.01 & 0.06*** & -0.01 & -0.01 & 0.02 & 0.00 & 0.03** & -0.04*** \\
			\bottomrule
		\end{tabular}
		
		\vspace{1mm}
		\begin{tablenotes}
			\footnotesize
			\item \textbf{Note:} ***p$<0.001$; **p$<0.05$.
		\end{tablenotes}
	\end{table}
	
\end{landscape}

As shown in Figure \ref{fig:10}(a) and (b), We have plotted scatter diagrams of the reviewers' scores against the mentioned aspects in both ICLR and NeurIPS. ICLR and NeurIPS show similar overall results. From the figure, we can observe that, the mention of summary has a slightly positive influence on the assigned score, though the effect is very minimal. Motivation follows a similar pattern, showing a weak positive correlation with the score, but the scattered distribution of data points indicates that the impact is not significant. The regression lines for substance, soundness, and clarity are nearly horizontal, suggesting no significant relationship with the score. Mentions of replicability and meaningful comparison may have a weak negative impact on the score. In contrast, originality shows a very slight positive correlation, implying that this aspect could contribute to a higher score. We note that the regression lines are included for visualization purposes only and are not used for formal statistical inference.\\
\indent Furthermore, we calculated the correlation between the aspects mentioned and the scores assigned by reviewers in ICLR and NeurIPS, as shown in Table \ref{tab9}. As shown in the results from the table, consistent with the scatter plot findings, aspects such as summary, motivation, and originality are positively correlated with the assigned scores, while aspects like replicability and meaningful comparison exhibit a negative correlation with the scores. These results indicate that the number of mentions of various aspects has a weak influence on the assigned scores, with the correlations being not statistically significant. This suggests that even with LLM-assisted reviews, the use of LLM does not significantly impact the reviewers' assigned scores.
\subsubsection{Correlation Analysis Between Aspect Mentions and Reviewer-Assigned Confidence Scores}
\begin{figure}[H]%
	\centering
	\includegraphics[width=1\textwidth]{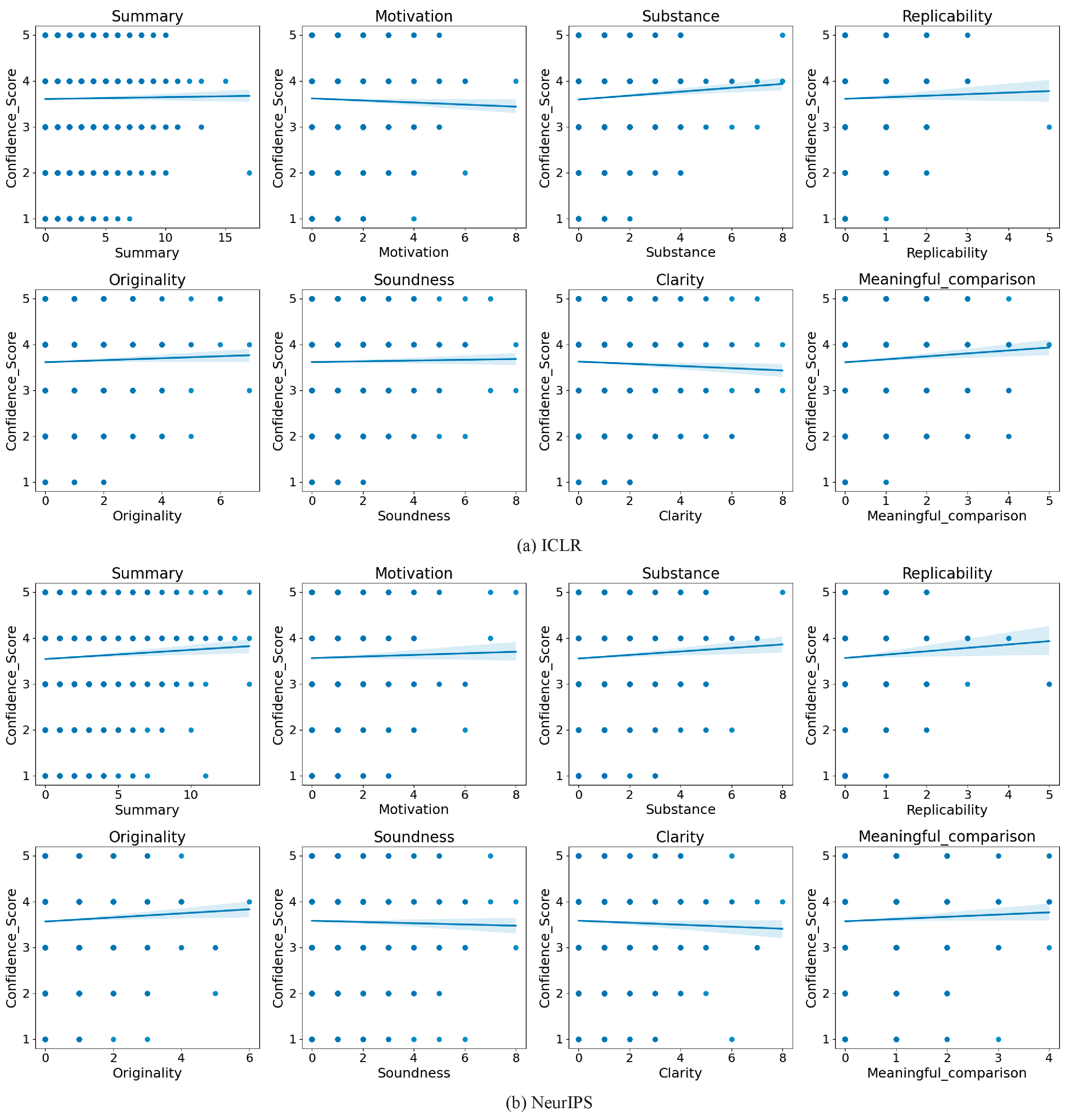}
	\caption{\centering{Scatter plots of confidence score versus aspects mentioned in ICLR and NeurIPS.}
	\textbf{Note:} The solid lines represent simple linear regression fits and the shaded areas show the associated confidence intervals. The figure is intended as a visual aid for qualitative comparison rather than as the basis for formal regression analysis.
	}
	\label{fig:11}
\end{figure}

We think that the use of LLM assistance may also have some influence on the confidence scores. Therefore, we also plotted scatter diagrams of the reviewers' confidence scores against the mentioned aspects in both ICLR and NeurIPS, shown in Figure \ref{fig:11}(a) and (b). ICLR and NeurIPS show similar overall results. 
From the figure, we can observe that, the relationship between aspect mention and reviewer confidence scores is generally weak in both ICLR and NeurIPS. While some aspects (e.g., abstract, substance, originality, and meaningful comparisons) show a slight positive correlation with confidence scores, these trends are accompanied by significant dispersion of data points. In contrast, clarity consistently shows a weak negative correlation with confidence scores. Overall, the scatter plots do not reveal a clear or robust linear relationship between aspect mention and confidence scores. We note that the regression lines are included for visualization purposes only and are not used for formal statistical inference.\\
\indent To quantitatively assess these observations, we calculated the Spearman correlation coefficient between aspect mention and confidence scores, as shown in Table \ref{tab10}. The results indicate that most correlations are weak in both conferences. Although some coefficients reached statistical significance, their values remain low, suggesting limited actual correlations. In particular, clarity shows a persistent negative correlation with confidence scores in both ICLR and NeurIPS, while most other arguments show a weak positive correlation. \\
\indent Overall, the association between the frequency of aspect mentions and reviewers' confidence scores is limited, indicating that reviewers' confidence under LLM assistance is not necessarily driven primarily by the relative emphasis they place on individual review aspects.
\begin{landscape}
	\begin{table}[t]
		\vspace{-0.8cm}
		\large
		\caption{\centering{The results of correlation analysis between aspect mentions and confidence scores in ICLR and NeurIPS. Summ represents Summary, Moti represents Motivation, Rep represents Replicability, Ori represents Originality, Meaningful represents Meaningful comparison.}}\label{tab10}%
		\centering
		\begin{tabular}{@{}m{15mm}<{\centering} m{15mm}<{\centering} m{18mm}<{\centering} m{18mm}<{\centering} m{15mm}<{\centering} m{18mm}<{\centering} m{18mm}<{\centering} m{18mm}<{\centering} m{15mm}<{\centering} m{20mm}<{\centering}@{}}
			\toprule
			\textbf{Venue} & & \textbf{Summ}&\textbf{Moti}&\textbf{Substance}&\textbf{Rep}&\textbf{Ori}&\textbf{Soundness}&\textbf{Clarity}&\textbf{Meaningful} \\
			\midrule
			ICLR&	Confidence score&	0.01&	-0.02**&	0.04***&	0.01&	0.02**&	0.00&	-0.02**&	0.03***\\
			NeurIPS&	Confidence score&	0.03**&	0.01***&	0.03**&	0.02**&	0.03**&	-0.01&	-0.02**&	0.02**\\
			
			\bottomrule 
		\end{tabular}
		\begin{tablenotes}
			\footnotesize
			\item \textbf{Note:} ***p$\textless$0.001; **p $\textless$ 0.05.
		\end{tablenotes}
	\end{table}
\end{landscape}
\section{Discussion}
In this section, we will discuss the implication of our study on theoretical and practical, and limitation of our study.
\subsection{Implication}
\textbf{(1) Theoretical Implication}\\
This study used peer review reports from two artificial intelligence conferences as data sources, collecting peer review report data from ICLR 2017-2025 and NeurIPS 2016-2024 (excluding 2020). We conducted aspect identification at the sentence level and performed word-level, sentence-level, and aspect-level analyses of these reports, both holistically and across different confidence scores. Furthermore, we also identified instances of LLM-assisted writing in the peer review reports for ICLR 2024-2025 and NeurIPS 2023-2024, and conducted a fine-grained analysis based on these findings. Based on the above results, we present the following insights. \\
\indent First, from a more fine-grained perspective, our results indicate that the number of words and sentences in ICLR and NeurIPS reviews have been affected to varying degrees following the emergence of LLMs. Similarly, with the exception of summary, the frequency of aspect mentions aligns with the observed trends in text length. Moreover, our findings reveal a decline in lexical complexity accompanied by an increase in the use of nominal subjects after the introduction of LLMs. These observations are consistent with the findings reported by Liang et al. \cite{r19} and Latona et al. \cite{r20}, their results indicate that at least 15\% of review reports were produced with the assistance of artificial intelligence, and we believe that this proportion will continue to increase. Therefore, we argue that the rapid advancement of artificial intelligence in recent years has influenced the way peer review reports are written.\\
\indent Secondly, we performed aspect identification on LLM-assisted and non-LLM-assisted review reports from ICLR 2024-2025 and NeurIPS 2023-2024, comparing the distribution of aspect mentions between the two groups. The results show that in ICLR, aside from the summary, the top three aspects emphasized in non-LLM-assisted reports were clarity, soundness, and originality, while in LLM-assisted reports, the focus shifted to clarity, soundness, and substance, accompanied by a noticeable reduction in originality-related mentions. In contrast, NeurIPS exhibits no significant changes across these aspects in our dataset, likely because the reviews primarily from accepted papers. This suggests that reviewers using LLM assistance may lack a clear framework for assessing originality, leading to fewer mentions of this aspect and placing more emphasis on others. \\
\indent Finally, we examine the relationship between aspect mentions in LLM-assisted review reports and reviewers' assigned scores. Both the scatter plots and Spearman correlation results indicate that aspect mentions exhibit only weak and largely insignificant correlations with overall scores, suggesting that LLM assistance does not substantially influence the final evaluations assigned to papers. This finding is encouraging, as it implies that reviewers primarily leverage LLMs to support the writing and organization of review reports, rather than to directly determine evaluative judgments. We further analyze the correlation between aspect mentions and reviewers' confidence scores. While most aspects show weak positive correlations with confidence, clarity consistently exhibits a negative correlation across both conferences. Although the correlations are weak, this result allows us to speculate that LLM assistance may not necessarily improve reviewers' understanding of the paper content. On the contrary, when clarity is low, reviewers may experience greater uncertainty even with LLM support, which can in turn reduce their confidence in their evaluations.\\
\indent In conclusion, these findings contribute to a nuanced theoretical understanding of how LLM assistance can reshape traditional peer review processes by altering the structure and focus of review content without directly modifying evaluation outcomes. These insights underscore the need for further research in developing LLM that not only improve efficiency but also enhance comprehension in expert review settings. It is essential to clarify that we do not intend to suggest that using LLM in review writing is inherently beneficial or detrimental. Nor do we claim (nor do we believe) that many reviewers are using LLM to compose entire reviews directly. Rather, we posit that current peer review processes are unlikely to be replaced by LLM but can instead be meaningfully supported by it.\\
\textbf{(2) Practical Implication}\\
Artificial intelligence is not a looming threat, and we need to approach its impact on academia with a rational perspective. While some studies \cite{r16,r17} have raised concerns that LLMs (such as ChatGPT) may hinder the peer review process, the ultimate control still lies in human hands. At present, LLM is merely a tool for assistance. Recent studies \cite{r26,r27,r28,r29} have also shown that LLMs have the potential to support peer review. We believe that appropriate guidelines or regulations should be established to restrict the use of LLMs in peer review, without completely prohibiting their use. Instead, journals or conferences should provide peer reviewers with LLM-assisted tools, rather than leaving the choice entirely to the reviewers. We are confident that the integration of LLM will significantly alleviate the pressure caused by the increasing volume of submissions. Recently, ICLR 2025 introducing\footnote{\href{https://blog.iclr.cc/2024/10/09/iclr2025-assisting-reviewers/}{https://blog.iclr.cc/2024/10/09/iclr2025-assisting-reviewers/}} a review feedback agent that identifies potential issues in reviews and provides feedback to reviewers for improvements. The goal of this system is to help make reviews more constructive and actionable for authors. From our results, the implementation of this system appears to have yielded a certain positive effect.\\
\indent Based on our findings, LLMs have not yet had a significant impact on the peer review process, particularly when examined at a fine-grained level. Reviewers using LLM assistance tend to employ it primarily for summarization or improving clarity rather than for evaluating specific aspects of a paper, such as soundness. While the study by Latona et al. \cite{r20} suggests that LLM-assisted reviews tend to assign higher scores, our fine-grained analysis reveals both positive and negative correlations between the use of LLM and the scores assigned to papers. This means that LLM assistance may lead to higher or lower scores, though overall, positive correlations are more common, aligning with Latona et al.'s findings. Therefore, we believe that the judicious use of LLM in tasks such as language polishing or summarizing paper content has a relatively minor impact on the review process. As previously discussed, certain LLM functionalities should be restricted until the technology has fully matured, which remains a long-term objective. 
\subsection{Discussion on the Impact of LLMs on Review Report}
While our analysis primarily focuses on linguistic and structural shifts in review texts, an important open question remains: whether the involvement of LLMs leads to more accurate, insightful, or useful reviews. Quantifying such dimensions of review quality is inherently challenging, as they involve subjective and context-dependent judgments that go beyond surface-level linguistic features. Nevertheless, our findings offer indirect evidence that may inform this broader discussion.\\
\indent The increase in clarity and soundness mentions among LLM-assisted reviews suggests that reviewers aided by LLMs are able to produce more coherent and logically consistent feedback. This linguistic improvement could enhance the readability and interpretability of review reports, potentially making them more accessible and actionable for authors. However, we also observe a decrease in the frequency of originality and meaningful comparison mentions-dimensions that are crucial for deep, evaluative reasoning. This may imply that while LLMs help reviewers articulate their ideas more fluently, they might also encourage reliance on generic or template-like expressions, thereby reducing the depth and specificity of critical engagement.\\
\indent In other words, LLM assistance appears to improve the form and presentation of reviews rather than their analytical rigor. Future work could investigate this issue more directly by combining linguistic analyses with expert evaluations of review helpfulness, accuracy, and constructiveness. Such a multi-dimensional assessment-integrating human judgments, author feedback, and content-based scoring-would provide a more definitive understanding of whether LLMs enhance not only how reviewers write but also how well they evaluate.
\subsection{Limitation}
Our study has several limitations worth highlighting. First, our data is primarily drawn from top conferences in the fields of machine learning and deep learning. To generalize these findings to other fields, corresponding data from those areas would be required. The NeurIPS data we used primarily comprises accepted papers, while the ICLR data contains more rejected papers than accepted ones. This difference may influence certain results, such as average length, aspect mentions, and potential LLM assistance usage. Additionally, our research conducts fine-grained analysis mainly through word and sentence counts as well as lexical and syntactic complexity. We did not further examine detailed word usage or conduct more in-depth syntactic analyses, such as changes in syntactic dependencies. Secondly, the LLM-assisted detection method we used identifies reports that may have LLM assistance but does not provide absolute certainty, as we enhance detection through likely LLM-assisted lexicons. Finally, our study is based on analyses of existing correlations or observed patterns rather than aiming to infer causal relationships through experimental or quasi-experimental methods. Finally, this study focuses on identifying correlations and observed patterns rather than establishing causation. Although our analyses highlight associations between review text, review aspect, sentiment, overall scores, confidence scores, and LLM-assisted, they do not imply causal relationships due to the observational nature of our study. Experimental or quasi-experimental approaches, such as controlled studies of LLM's impact on review writing, would be necessary to draw causal inferences.
\section{Conclusion and Future Works}
In this paper, we conducted a fine-grained analysis of open peer review reports from AI conferences in recent years. First, we examined the overall changes in the length of review texts, the distribution of aspect mentions, and the sentiment polarity of different aspects over time. We then performed the same analysis on review reports with varying levels of confidence scores. Additionally, we analyzed the lexical and syntactic complexity of peer review texts in recent years. Then, we analyzed the aspect mentions in LLM-assisted and non-LLM-assisted review reports. Finally, we calculated the correlation between the assigned scores and aspect mentions in LLM-assisted review reports. Overall, our findings indicate that the emergence of LLMs is associated with longer and more fluent peer-review texts, increased emphasis on summaries and surface-level clarity, and increasingly standardized linguistic patterns effects that are particularly pronounced among reviewers with lower self-reported confidence. At the same time, attention to deeper evaluative dimensions, including originality, replicability, and fine-grained critical reasoning, has declined. These shifts become more evident when comparing LLM-assisted and non-LLM-assisted reviews. Although LLM-assisted reports show a modest improvement in the informativeness of recommendations, this trend raises concerns about a potential trade-off between linguistic fluency and evaluative depth in peer review. We believe that our approach possesses a degree of generalizability across different domains and is not limited to the specific fields associated with the data discussed in this study.\\
\indent In future work, we plan to explore the causal relationship between LLM usage and changes in peer review comments by employing experimental or quasi-experimental methodologies. Additionally, we intend to broaden our scope by analyzing peer review reports from leading journals such as PLOS ONE and Nature Communications, to identify trends and patterns in the review process across different fields and publication standards. Finally, we aim to adopt more sophisticated tools and techniques for detecting the use of LLMs in peer review comments. These tools could include advanced text analysis methods and LLM-based detection systems, which will help us identify and quantify the extent to which LLMs influence scholarly reviews and ensure transparency in the review process.

\bmhead{Acknowledgements}
\noindent This study is supported by the National Natural Science Foundation of China (Grant No. 72074113).

\section*{Declarations}
The author(s) declared no potential conlicts of interest with respect to the research, author- ship, and/or publication of this article.

\begin{appendices}

\section{Distribution of Review Scores and Confidence Scores}
\label{app2}
\setcounter{table}{0}
\renewcommand{\thetable}{A\arabic{table}}
Table~\ref{tab:taba2} and ~\ref{tab:taba3} are distribution of review scores and confidence scores in ICLR and NeurIPS.

\begin{table}[htbp]
	\centering
	\caption{Distribution of review scores in ICLR 2017-2025 and NeurIPS 2016-2024.}
	\label{tab:taba2}
	\begin{tabular}{@{}ccccccccccc@{}}
		\hline
		Year & 1 & 2 & 3 & 4 & 5 & 6 & 7 & 8 & 9 & 10 \\ \hline
		ICLR 2017 & 1 & 14 & 251 & 258 & 338 & 331 & 138 & 930 & 36 & 6 \\
		ICLR 2018 & 5 & 60 & 224 & 536 & 560 & 621 & 523 & 170 & 	47 & 2 \\
		ICLR 2019 & 15 & 91 & 418 & 983 & 1063 & 1067 & 833 & 222 & 66 & 6 \\
		ICLR 2020 & 924 & - & 2562 & - & - & 2393 & - & 842 & - & - \\
		ICLR 2021 & 14 & 167 & 905 & 2289 & 2685 & 2832 & 1999 & 495 & 112 & 10 \\
		ICLR 2022 & 256 & - & 2769 & - & 3224 & 2972 & - & 3508 & - & 48 \\
		ICLR 2023 & 465 & - & 4704 & - & 5121 & 5324 & - & 2845 & - & 101\\
		ICLR 2024 & 353 & - & 4246 & - & 6153 & 7632 & - & 3759 & - & 102 \\
		ICLR 2025 & 941 & - & 10506 & - & 11547 & 13457 & - & 5950 & - & 170 \\
		\hline
		NeurIPS 2016 & 941 & - & 10506 & - & 11547 & 13457 & - & 5950 & - & 170 \\
		NeurIPS 2017 & 941 & - & 10506 & - & 11547 & 13457 & - & 5950 & - & 170 \\
		NeurIPS 2018 & 941 & - & 10506 & - & 11547 & 13457 & - & 5950 & - & 170 \\
		NeurIPS 2019 & 941 & - & 10506 & - & 11547 & 13457 & - & 5950 & - & 170 \\
		NeurIPS 2021 & 2 & 18 & 140 & 513 & 1259 & 3939 & 3788 & 905 & 156 & 9 \\
		NeurIPS 2022 & 4 & 37 & 320 & 836 & 1935 & 3445 & 2913 & 767 & 69 & 4 \\
		NeurIPS 2023 & 8 & 50 & 477 & 1197 & 3468 & 5075 & 3933 & 887 & 66 & 10 \\
		NeurIPS 2024 & 8 & 45 & 506 & 1316 & 4163 & 5433 & 4140 & 930 & 80 & 14 \\
		\hline
	\end{tabular}

\end{table}
\begin{table}[htbp]
	\centering
	\caption{Distribution of confidence scores in ICLR 2017-2025 and NeurIPS 2021-2024.}
	\label{tab:taba3}
	\begin{tabular}{@{}ccccccccccc@{}}
		\hline
		Year & 1 & 2 & 3 & 4 & 5  \\ \hline
		ICLR 2017 & 7 & 51 & 387 & 804 & 249  \\
		ICLR 2018 & 34 & 130 & 692 & 1391 & 501 \\
		ICLR 2019 & 40 & 275 & 1239 & 2378 & 832  \\
		ICLR 2020 & - & - & - & - & -  \\
		ICLR 2021 &92 & 641 & 3327 & 5614 & 1834  \\
		ICLR 2022 & 17 & 652 & 4223 & 6521 & 1364  \\
		ICLR 2023 & 87 & 1364 & 6077 & 8922 & 2110  \\
		ICLR 2024 & 113 & 1736 & 7639 & 10226 & 2531  \\
		ICLR 2025 & 149 & 2787 & 13842 & 20181 & 5612  \\
		\hline
		NeurIPS 2021 & 100 & 674 & 3578 & 5182 & 1195 \\
		NeurIPS 2022 & 166 & 855 & 3582 & 4625 & 1102 \\
		NeurIPS 2023 & 271 & 1315 & 5231 & 6669 & 1685 \\
		NeurIPS 2024 & 201 & 1326 & 5391 & 7643 & 2074  \\
		\hline
	\end{tabular}
\end{table}

\section{Lexicon of Frequently and Predominantly Used Terms by LLMs}
\label{app1}
\setcounter{table}{0}
\renewcommand{\thetable}{B\arabic{table}}
Table~\ref{taba1} is the lexicon of frequently and predominantly used terms by LLMs.

\begin{longtable}{@{}lll@{}}
	\caption{\centering Lexicon of frequently and predominantly used terms by LLMs}
	\label{taba1}\\
	\toprule
	meticulously & reportedly & lucidly \\
	innovatively & aptly & methodically \\
	excellently & compellingly & impressively \\
	undoubtedly & scholarly & strategically \\
	intriguingly & competently & intelligently \\
	hitherto & thoughtfully & profoundly \\
	undeniably & admirably & creatively \\
	logically & markedly & thereby \\
	contextually & distinctly & judiciously \\
	cleverly & invariably & successfully \\
	chiefly & refreshingly & constructively \\
	inadvertently & effectively & intellectually \\
	rightly & convincingly & comprehensively \\
	seamlessly & predominantly & coherently \\
	evidently & notably & professionally \\
	subtly & synergistically & productively \\
	purportedly & remarkably & traditionally \\
	starkly & promptly & richly \\
	nonetheless & elegantly & smartly \\
	solidly & inadequately & effortlessly \\
	forth & firmly & autonomously \\
	duly & critically & immensely \\
	beautifully & maliciously & finely \\
	succinctly & further & robustly \\
	decidedly & conclusively & diversely \\
	exceptionally & concurrently & appreciably \\
	methodologically & universally & thoroughly \\
	soundly & particularly & elaborately \\
	uniquely & neatly & definitively \\
	substantively & usefully & adversely \\
	primarily & principally & discriminatively \\
	efficiently & scientifically & alike \\
	herein & additionally & subsequently \\
	potentially & commendable & innovative \\
	meticulous & intricate & notable \\
	versatile & noteworthy & invaluable \\
	pivotal & potent & fresh \\
	ingenious & cogent & ongoing \\
	tangible & profound & methodical \\
	laudable & lucid & appreciable \\
	fascinating & adaptable & admirable \\
	refreshing & proficient & intriguing \\
	thoughtful & credible & exceptional \\
	digestible & prevalent & interpretative \\
	remarkable & seamless & economical \\
	proactive & interdisciplinary & sustainable \\
	optimizable & comprehensive & vital \\
	pragmatic & comprehensible & unique \\
	fuller & authentic & foundational \\
	distinctive & pertinent & valuable \\
	invasive & speedy & inherent \\
	considerable & holistic & insightful \\
	operational & substantial & compelling \\
	technological & beneficial & excellent \\
	keen & cultural & unauthorized \\
	strategic & expansive & prospective \\
	vivid & consequential & manageable \\
	unprecedented & inclusive & asymmetrical \\
	cohesive & replicable & quicker \\
	defensive & wider & imaginative \\
	traditional & competent & contentious \\
	widespread & environmental & instrumental \\
	substantive & creative & academic \\
	sizeable & extant & demonstrable \\
	prudent & practicable & signatory \\
	continental & unnoticed & automotive \\
	minimalistic & intelligent & underscores \\
	necessitating & delves & adaptability \\
	delved & delve & elucidated \\
	underscore & credibility & advancements \\
	elucidation & underpinnings & equitable \\
	perplexing & excels & intricacies \\
	persuasiveness & delineation & elucidate \\
	provision & bolster & discourse \\
	meticulous & endeavors & tangible \\
	commendable & showcasing & imperative \\
	encompassing & offering &  \\
	\bottomrule
	
\end{longtable}
\end{appendices}


\bibliography{sn-bibliography}

\end{document}